\documentclass[10pt,twocolumn,letterpaper]{article}

\usepackage{wacv}
\usepackage{times}
\usepackage{epsfig}
\usepackage{graphicx}
\usepackage{amsmath}
\usepackage{amssymb}
\usepackage[table,xcdraw]{xcolor}
\usepackage{multirow}
\usepackage{color}
\usepackage{arydshln}
\usepackage{float}
\usepackage{caption}
\def\wacvPaperID{456}

\wacvfinalcopy % *** Uncomment this line for the final submission

\ifwacvfinal
\def\assignedStartPage{9876} 
\fi

\def\smoothingloss{$\mathcal{L}_s$}

\def\first{\cellcolor[HTML]{FFFFBB}}
\def\third{\cellcolor[HTML]{FFCCC9}}
\def\second{\cellcolor[HTML]{DEFFC9}}

\ifwacvfinal
\usepackage[breaklinks=true,bookmarks=false]{hyperref}
\else
\usepackage[pagebackref=true,breaklinks=true,colorlinks,bookmarks=false]{hyperref}
\fi

\ifwacvfinal
\else
\fi

\begin{document}

%%%%%%%%% TITLE
\title{Auto White-Balance Correction for Mixed-Illuminant Scenes}

\author{\hspace{7mm} Mahmoud Afifi$^{1}$ \hspace{10mm} Marcus A. Brubaker$^{1, 2}$ \hspace{10mm} Michael S. Brown$^{1, 2}$\hspace{7mm}\vspace{3mm}\\
$^{1}$York University \hspace{15mm} $^{2}$Vector Institute\vspace{2mm}\\
{\tt\small \{mafifi,mab,mbrown\}@eecs.yorku.ca}}

\twocolumn[{%
\renewcommand\twocolumn[1][]{#1}%
\maketitle
\begin{center}
\includegraphics[width=\textwidth]{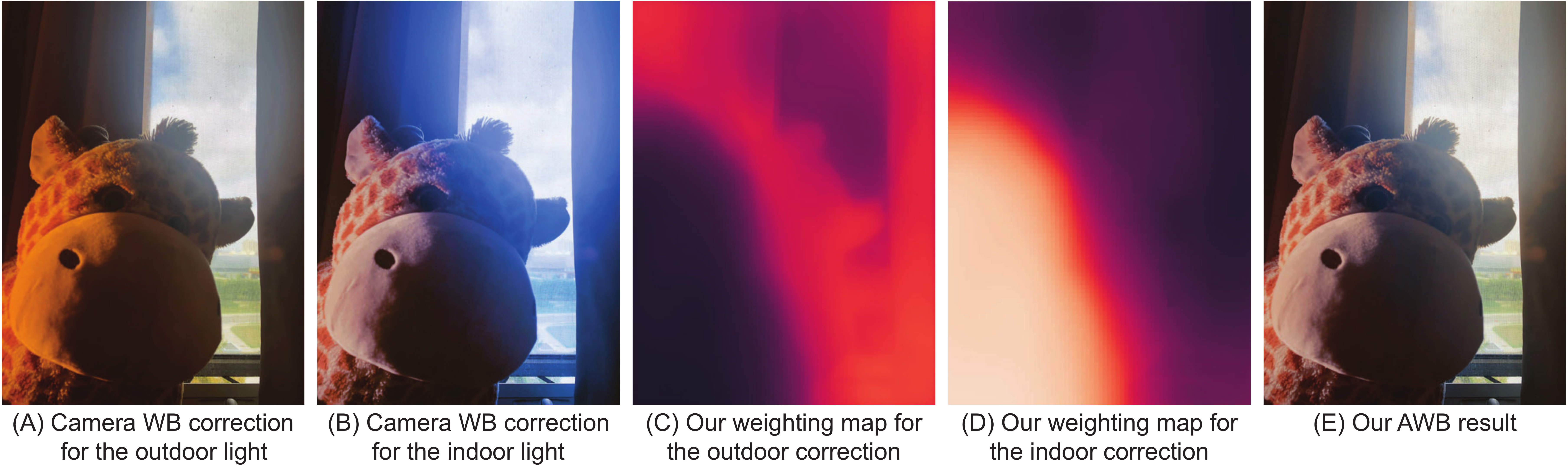}
\vspace{-6mm}
\captionof{figure}{Camera AWB correction corrects colors for only a single illumination present in a captured scene. In the mixed-illuminant scenario, this traditional WB correction often results in undesirable color tint (typically, reddish or bluish) in the final sRGB image, as shown in (A) and (B). Our method can effectively deal with such mixed-illuminant scenes by learning a set of local weighting maps to blend between different WB settings, as shown in (C) and (D), producing more compelling AWB correction compared to traditional camera AWB, as shown in (E).}
\label{fig:teaser}
\end{center}%
}]

\maketitle
\thispagestyle{empty}

%%%%%%%%% ABSTRACT
\begin{abstract}
\vspace{-2mm}
Auto white balance (AWB) is applied by camera hardware at capture time to remove the color cast caused by the scene illumination.  The vast majority of white-balance algorithms assume a single light source illuminates the scene; however, real scenes often have mixed lighting conditions. This paper presents an effective AWB method to deal with such mixed-illuminant scenes. A unique departure from conventional AWB, our method does not require illuminant estimation, as is the case in traditional camera AWB modules. Instead, our method proposes to render the captured scene with a small set of predefined white-balance settings. Given this set of rendered images, our method learns to estimate weighting maps that are used to blend the rendered images to generate the final corrected image. Through extensive experiments, we show this proposed method produces promising results compared to other alternatives for single- and mixed-illuminant scene color correction.  Our source code and trained models are available at \href{https://github.com/mahmoudnafifi/mixedillWB}{https://github.com/mahmoudnafifi/mixedillWB}. 
\end{abstract}

%%%%%%%%% BODY TEXT
\section{Introduction} \label{sec:intro}
Auto white balance (AWB) is an essential procedure applied by a camera's dedicated image signal processor (ISP) hardware. The ISP ``renders'' the sensor image to the final output through a series of processing steps, such as white balance (WB), tone-mapping, and conversion of the final standard RGB (sRGB) output image.  WB is performed early in the ISP pipeline and aims to remove undesirable color casts caused by the environment lighting~\cite{ebner2007color}. AWB mimics the human visual system's ability to perform color constancy, which allows us to perceive objects as the same color regardless of scene lighting.

AWB consists of two steps. First, the scene illuminant color as observed by the camera's sensor is estimated using an illuminant estimation algorithm (e.g., \cite{DSNET, hu2017fc, ccc, ffcc}). %, c5}). 
This first step assumes there is a single (i.e., global) illuminant in the scene.  Second, the captured image is corrected based on the estimated illumination. These two steps are applied early in the camera's ISP to the raw image \cite{gehler2008bayesian}.

The majority of prior work focuses on the illuminant estimation step, while the WB procedure is performed using a simple diagonal-based correction process \cite{gijsenij2011computational}. Global illumination estimation is a well-studied topic and recent work (e.g., \cite{xu2020end, hernandez2020multi, lo2021clcc}) achieves impressive results of less than 2\textdegree $ $ angular errors\footnote{Perceptually, angular errors less than 2\textdegree $ $ are considered barely noticeable.} across all available single-illuminant datasets. 

The assumption of a single global illuminant is well known to be an oversimplification of the problem, as many captured scenes have multiple light sources.  Performing single-illuminant AWB on scenes with multiple illuminations often produces unsatisfactory results \cite{bleier2011color}. Figure \ref{fig:teaser} shows a typical example of a scene captured with a mixed lighting condition. The colors of the captured scene are corrected by the camera AWB module, which applies a single-illuminant WB correction after running an illuminant estimation algorithm on the ISP. The white-balanced image is then processed through a set of color rendering functions typical of an ISP to produce the final sRGB image, shown in Figure \ref{fig:teaser}-(A). Because the captured scene has two different light sources -- outdoor lighting on the right side of the scene and indoor lighting on the left side of the scene -- traditional camera AWB correction results in either a reddish or bluish tint in the final rendered image.  

\paragraph{Contribution}
In this paper, we present an AWB method to deal with both single- and mixed-illuminant scenes. Unlike traditional camera AWB modules, our method \textit{does not} require illuminant estimation; instead, we propose to render the captured scene multiple times with a few fixed WB settings. Given these rendered images of the same scene, we outline a method to learn suitable pixel-wise blending maps to generate the final sRGB image. Our generated spatially varying weighting maps allow us to correct for different lighting conditions in the captured scene, as shown in Figure \ref{fig:teaser}-(C--E). As a part of this effort, we propose a synthetic test set of mixed-illuminant scenes with pixel-wise ground truth. We use this set, along with other datasets (\cite{banic2017unsupervised, fivek}), to validate our method through an extensive set of experiments and comparisons. 

\section{Related Work} \label{sec:related-work}

\begin{figure*}[t]
\centering
\includegraphics[width=\linewidth]{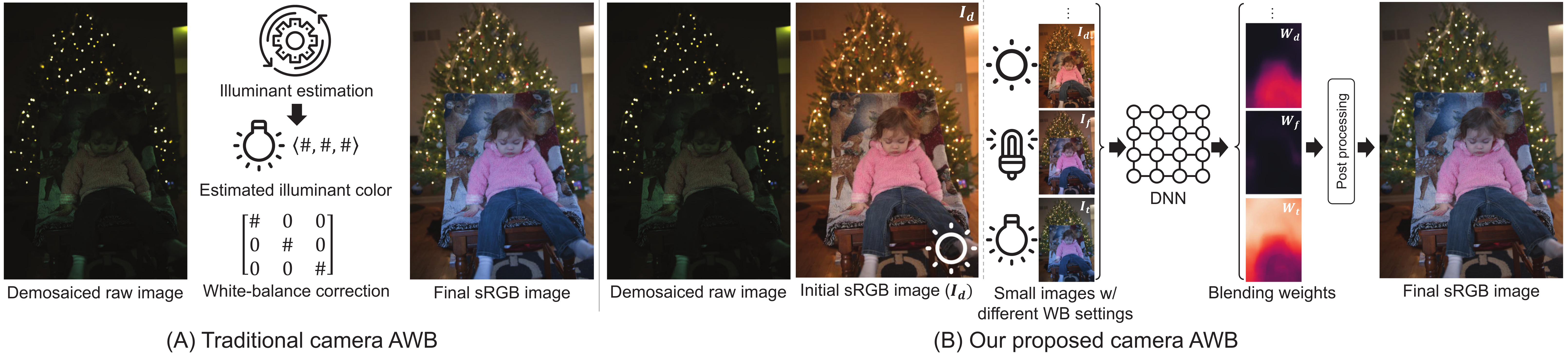}
\vspace{-6mm}
\caption{Proposed AWB method. Unlike  traditional camera AWB, shown in (A), which corrects colors for only a single illuminant estimated by an illuminant estimation method, our AWB method, shown in (B), applies a single WB correction that corrects for a fixed color temperature (e.g., 5500 K) in addition to generating small images ($384\!\times\!384$) with a predefined set of color temperatures (e.g., 2850K, 3800K) that correlate to common lighting conditions (e.g., incandescent, fluorescent). Given these images, our method predicts blending weights to produce our final AWB result. \label{fig:method}}
\end{figure*}

This section briefly reviews prior work for illuminant estimation and WB correction for sRGB images. 

\subsection{Illuminant Estimation} \label{subsec:ill-estimation}
Working directly on a raw sensor image, illuminant estimation methods aim to predict the global scene illumination color \cite{gijsenij2011computational}. A large body of literature performs this estimation according to some statistical hypothesis on the image colors/edges. A common statistical hypothesis for illuminant estimation is the gray-world hypothesis \cite{GW}, which assumes the mean of image irradiance is achromatic (i.e., ``gray'') and thus computing the mean image colors could give a rough estimation of the scene illuminant color. Other statistical methods include the white-patch hypothesis \cite{maxRGB}, the bright pixels \cite{BP}, the shades-of-gray \cite{SoG}, the gray-edges \cite{GE}, the weighted-gray-edges \cite{gijsenij2012improving}, the bright-and-dark colors PCA \cite{cheng2014illuminant}, the gray pixel \cite{GP}, and the grayness index \cite{GI}. 

Despite being hardware-friendly, the accuracy of such statistical methods is not always satisfying. Learning-based methods offer more accurate results by relying on training examples to predict illuminant colors at inference time. Representative examples of learning-based illuminant estimation methods include: gamut-based methods \cite{forsyth1990novel, finlayson2000improving, Gamut, Gamut1, bianco2012color},  Bayesian methods \cite{brainard1994bayesian, brainard1997bayesian, rosenberg2004bayesian, gehler2008bayesian}, bias correction methods \cite{MomentCorrection, royalsociety, afifi2019projective},
and neural network-based methods  \cite{funt1996learning, cardei2002estimating, BMVC1, DSNET, hu2017fc, Seoung, bianco2019quasi, xu2020end, hernandez2020multi, lo2021clcc, c5}.

While the majority of prior work adopts the single-illuminant assumption, only a handful of attempts have been proposed to deal with multi-illuminant scenes %to address this task 
(e.g., \cite{finlayson1995color, hsu2008light, gijsenij2011color, joze2013exemplar, beigpour2013multi, hussain2018color}). These prior methods employ pixel-wise illumination estimation strategies to address the problem.  In some cases, the number of different lights in the scene need to be known in advance.  In contrast to prior work, our AWB method can deal with single- and mixed-illuminant scenes \textit{without} an explicit illuminant estimation step on the camera imaging pipeline. 

\subsection{WB Correction for sRGB Images} \label{subsec:wb-correction}
As mentioned earlier, traditional camera ISPs use a simple diagonal correction to white-balance captured raw images, giving the estimated illuminant colors of the scenes. After white balancing, the camera ISP applies color rendering and photo-finishing steps to render the final sRGB image. If the AWB module had some errors during the camera ISP rendering -- which typically is the case when the captured scenes have mixed lighting conditions -- it becomes challenging to correct colors in post-capture. A few attempts were proposed recently to deal with such a scenario by replacing the diagonal correction with a non-linear correction function \cite{afifi2019color, afifi2019tuning, afifi2020interactive}. More recently, a deep neural network (DNN) was proposed to correct/edit the WB settings of camera-rendered sRGB images \cite{afifi2020deep}. 

These sRGB methods treat the problem as a global illumination correction post image capture. In our approach, we employ a modified camera ISP that allows us to benefit from the raw image early in the camera rendering chain. With this modified ISP, our method can effectively correct colors of both single- and mixed-illuminant scenes without requiring any post-capture correction (e.g., \cite{afifi2019color, afifi2020deep}) nor interaction from the user (e.g., \cite{afifi2019tuning, afifi2020interactive}).

\section{Method} \label{sec:method}

The overview of our method is depicted in Figure \ref{fig:method}. As shown, we adopted a modified camera ISP, originally introduced in \cite{afifi2019tuning}. This modified camera ISP produces additional small images of the same captured scene, each of which is rendered with a different WB setting.  Our approach works by observing the same scene object colors under a fixed set of WB settings.  The correct WB is then generated by linearly blending amoung the different WB settings. That is, given a set of small sRGB images $\{I_{{c_1}\downarrow}, I_{{c_2}\downarrow}, ...\}$, each of which was rendered with a different WB setting (i.e., $c_1$, $c_2$, ...), the final spatially varying WB colors are computed as follows:
\begin{equation}
I_{\texttt{corr}\downarrow} = \sum_{i} W_i \odot I_{{c_i}\downarrow},
\label{eq:formulation}
\end{equation}
\noindent where $I_{\texttt{corr}\downarrow}$ is the corrected small image, $i$ is for indexing, $\odot$ is Hadamard product, and $W_i \in \mathbb{R} ^{w\!\times\!h\!\times\!3}$ is the weighting map for the $i^\text{th}$ small image rendered with the $c_i$ WB setting. Here, $w$ and $h$ refer to the width and height of our small images, respectively. For simplicity, we can think of $W_i$ as a replicated  monochromatic image to match the dimensions of $I_i$. In our experiments at inference time, we used small images of $384\!\times\!384$ pixels (i.e., $w=h=384$).

Note that, we decided to learn the weighting maps, $\{W_i\}$, for the images rendered in the sRGB space -- after a full rendering of demosaiced raw images -- to make our method device-independent. With that said, our method can be easily adjusted to learn the weighting maps in the linear raw space. 

Producing $\{I_{{c_i}\downarrow}\}$ can be attained by following the modification introduced in \cite{afifi2019tuning} (Sec.\ \ref{subsec:cam-isp}). Then, our main task is to learn the values of $W_i$, given $\{I_{{c_i}\downarrow}\}$ (Sec.\ \ref{subsec:maps}). Lastly, we apply  post-processing steps to compute the final high-resolution white-balanced image (Sec.\ \ref{subsec:inference}).

\subsection{Modified Camera ISP} \label{subsec:cam-isp}
We followed the modified ISP proposed in \cite{afifi2019tuning}, which renders a few small images, each of which is rendered with one of a set of predefined WB settings, along with the high-resolution image rendered with the camera AWB correction. In contrast to \cite{afifi2019tuning}, we propose to use a fixed WB setting to render the high-resolution image, and thus we \textit{do not} need an illuminant estimation module in our pipeline. In the rest of this paper, the WB settings are interchangeably referred to by lighting source names or the correlated color temperatures. 

Rendering such small images through the camera ISP modules does not require major changes to standard camera ISPs. Specifically, this modification downsamples the captured raw demosaiced image into the target size of the small images. This small raw-image is then processed by the  ISP color rendering modules to produce a small sRGB image. For each rendering pass, a predefined WB setting is used. Processing such small images though the ISP an additional few times requires negligible processing time \cite{afifi2019tuning}, and thus it does not affect the real-time processing offered by the camera viewfinder. 

In the photofinishing stage, we can employ these small images to map the colors of the high-resolution sRGB image, rendered with a fixed WB setting (e.g., daylight), to our set of predefined WB settings as described in the following equation:

\begin{equation}
\hat{I}_{c_i} = M_{c_i} \phi\left(I_{\texttt{init}}\right),
\label{eq:mapping}
\end{equation}

\noindent where $I_{\texttt{init}}$ is the initial high-resolution image rendered with the fixed WB setting, $\hat{I}_{c_i}$ is the mapped image to the target WB setting $c_i$, $\phi\left(\cdot\right)$ is a polynomial kernel function that projects the R, G, B channels into a higher-dimensional space \cite{hong2001study}, and $M_{c_i}$ is a mapping matrix computed by fitting the colors of $I_{\texttt{init}}$, after downsampling, to the corresponding colors of $I_{{c_i}\downarrow}$. This fitting is computed by minimizing the residual sum of squares between the source and target colors in both small images.

\subsection{Learning Weighting Maps} \label{subsec:maps}

So far, we have described the rendering process of our small images and the post-capture color mapping to generate high-resolution images with our predefined set of WB settings. Our task now is to estimate the values of $W_i$ for a given set of our small images $\{I_{{c_i}\downarrow}\}$.

We employed a DNN to predict the values of $\{W_i\}$, where our network accepts the small images, rendered with our predefined WB settings, and learns to produce proper weighting maps $\{W_i\}$. Specifically, our network accepts a 3D tensor of concatenated small images and produces a 3D tensor of the weighting maps. The details of our network architecture are given in the appendix. 

To train our DNN model, we employed the Rendered WB dataset \cite{afifi2019color}, which includes $\sim$65K sRGB images rendered with different color temperatures. Each image in the Rendered WB dataset \cite{afifi2019color} has the corresponding ground-truth sRGB image that was rendered with an accurate WB correction. From this dataset, we selected 9,200 training images that were rendered with ``camera standard'' photofinishing and with the following color temperatures: 2850 Kelvin (K), 3800 K,
5500 K, 6500 K, and 7500 K, which correspond to the following WB settings: tungsten (also referred to as incandescent), fluorescent, daylight, cloudy, and shade, respectively \cite{afifi2019else}. 
We then train our network to minimize the following reconstruction loss function:
\begin{equation}
\mathcal{L}_r = \lVert P_\texttt{corr} - \sum_{i} \hat{W}_i \odot P_{c_i} \rVert_F^2,
\label{eq:rec_loss}
\end{equation}

\noindent where $ \lVert\cdot\rVert_F^2$ computes the squared Frobenius norm, $P_{\texttt{corr}}$ and $P_{c_i}$ are the extracted training patches from ground-truth sRGB images and input sRGB images rendered with the $c_i$ WB setting, respectively, and $\hat{W}_i$ is the blending weighting map produced by our network for $P_{c_i}$.

To avoid producing out-of-gamut colors in the reconstructed image, we apply a cross-channel softmax operator to the output tensor of our network before computing our loss in Equation \ref{eq:rec_loss}. 

We apply the following regularization term to the produced weighting maps to encourage our network to produce smooth weights:

\begin{equation}
\mathcal{L}_s = \sum_i \lVert \hat{W}_i \ast \nabla_x \rVert_F^2 + \lVert \hat{W}_i \ast \nabla_y \rVert_F^2  
\label{eq:smooth_loss}
\end{equation}

\noindent where $\nabla_x$ and $\nabla_y$ are $3\!\times\!3$ horizontal and vertical Sobel filters, respectively, and  $\ast$ is the convolution operator. Thus, our final loss is computed as follows:

\begin{equation}
\mathcal{L} = \mathcal{L}_r + \lambda \mathcal{L}_s, 
\label{eq:final_loss}
\end{equation}

\noindent where $\lambda$ is a scaling factor to control the contribution of \smoothingloss $ $ to our final loss. In our experiments, we used $\lambda=100$. Minimizing the loss in Equation \ref{eq:final_loss} was performed using the Adam optimizer \cite{kingma2014adam} with $\beta_1= 0.9$ and $\beta_2= 0.999$ for 
200 epochs. We used a mini-batch size of 32, where at each iteration we randomly selected eight training images, alongside their associated images rendered with our predefined WB set, and extracted four random patches from each of the selected training images. 

\subsection{Inference} \label{subsec:inference}

At inference time, we produce our small image with the predefined WB settings. These images are concatenated and fed to our DNN, which produces the weighting maps. To improve the final results, we advocate an ensembling strategy, where we feed the concatenated small images in three different scales: $1.0$, $0.5$, $0.25$. Then, we upsample the produced weights to the high-resolution image size. Afterward, we compute the average weighting maps produced for each WB setting. This ensemble strategy produces weights with more local coherence. This coherence can be further improved by applying a post-processing edge-aware smoothing step. That is, we used our high-resolution images (i.e., the initially rendered one with the fixed WB setting and the mapped images produced by Equation \ref{eq:mapping}) as a guidance image to post-process the generated weights. We utilized the fast bilateral solver for this task \cite{barron2016fast}. See the appendix for an ablation study. 

After generating our upsampled weighting maps $\{\hat{W}_{i\uparrow}\}$, the final image is generated as described in the following equation:
\begin{equation}
\hat{I}_\texttt{corr} = \sum_{i} \hat{W}_{i\uparrow} \odot \hat{I}_{c_i}.
\label{eq:formulation}
\end{equation}

\section{Experiments} \label{sec:experiments}

\begin{figure*}[t]
\centering
\includegraphics[width=\linewidth]{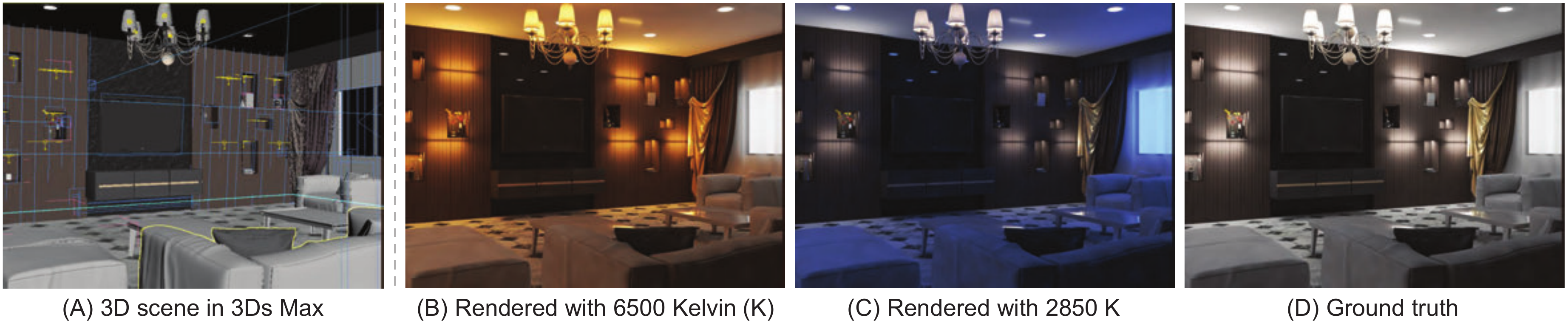}
\vspace{-6mm}
\caption{Our mixed-illuminant evaluation set. (A) We began with 3D scenes modeled in Autodesk 3Ds Max \cite{3dsmax}. (B-C) We set the WB setting of the virtual camera to different color temperatures and produced realistic scene images with multiple illuminations using Vray rendering \cite{vray}. (D) For each input image in our test set, a ground-truth image is provided. \label{fig:dataset}}
\end{figure*}

We conducted several experiments to evaluate our method. In all experiments, we set the fixed WB setting to 5500 K (i.e., daylight WB). We examined two different predefined WB sets to render our small images, which are: (i) : \{2850 K, 3800 K, 6500 K, 7500 K\} and (ii) \{2850 K, 7500 K\}. These two sets represent the following WB settings: \{tungsten, fluorescent, cloudy, and shade\} and \{tungsten, shade\}, respectively. Our network received the concatenated images, along with the downsampled daylight image. Thus, we will refer to the first predefined WB set as $\{\texttt{t},\texttt{f}, \texttt{d}, \texttt{c}, \texttt{s}\}$ and the second set is referred to as $\{\texttt{t}, \texttt{d}, \texttt{s}\}$, where the terms $\texttt{t}$, $\texttt{f}$, $\texttt{d}$, $\texttt{c}$, and $\texttt{s}$ refer to tungsten, fluorescent, daylight, cloudy, and shade, respectively. 

When WB=$\{\texttt{t}, \texttt{d}, \texttt{s}\}$ is used, our DNN takes $\sim$0.6 seconds on average without ensembling, while it takes $\sim$0.9 seconds on average when ensembling is used. This processing time is reported using a single NVIDIA GeForce GTX 1080 graphics card. When WB=$\{\texttt{t},\texttt{f}, \texttt{d}, \texttt{c}, \texttt{s}\}$ is used, our DNN takes $\sim$0.8 seconds and $\sim$0.95 seconds on average to process with and without ensembling, respectively. 

The average CPU processing time of the post-processing edge-aware smoothing for images with 2K resolution is $\sim$3.4 seconds and $\sim$5.7 seconds with WB=$\{\texttt{t}, \texttt{d}, \texttt{s}\}$ and WB=$\{\texttt{t},\texttt{f}, \texttt{d}, \texttt{c}, \texttt{s}\}$, respectively. This post-processing step is optional (see the appendix for comparisons) and its processing time can be reduced with a GPU implementation of the edge-aware smoothing step. 

In the remaining part of this section, we elaborate the details of our experiments and report both qualitative and quantitative results of our method.

\begin{figure*}[t]
\centering
\includegraphics[width=\linewidth]{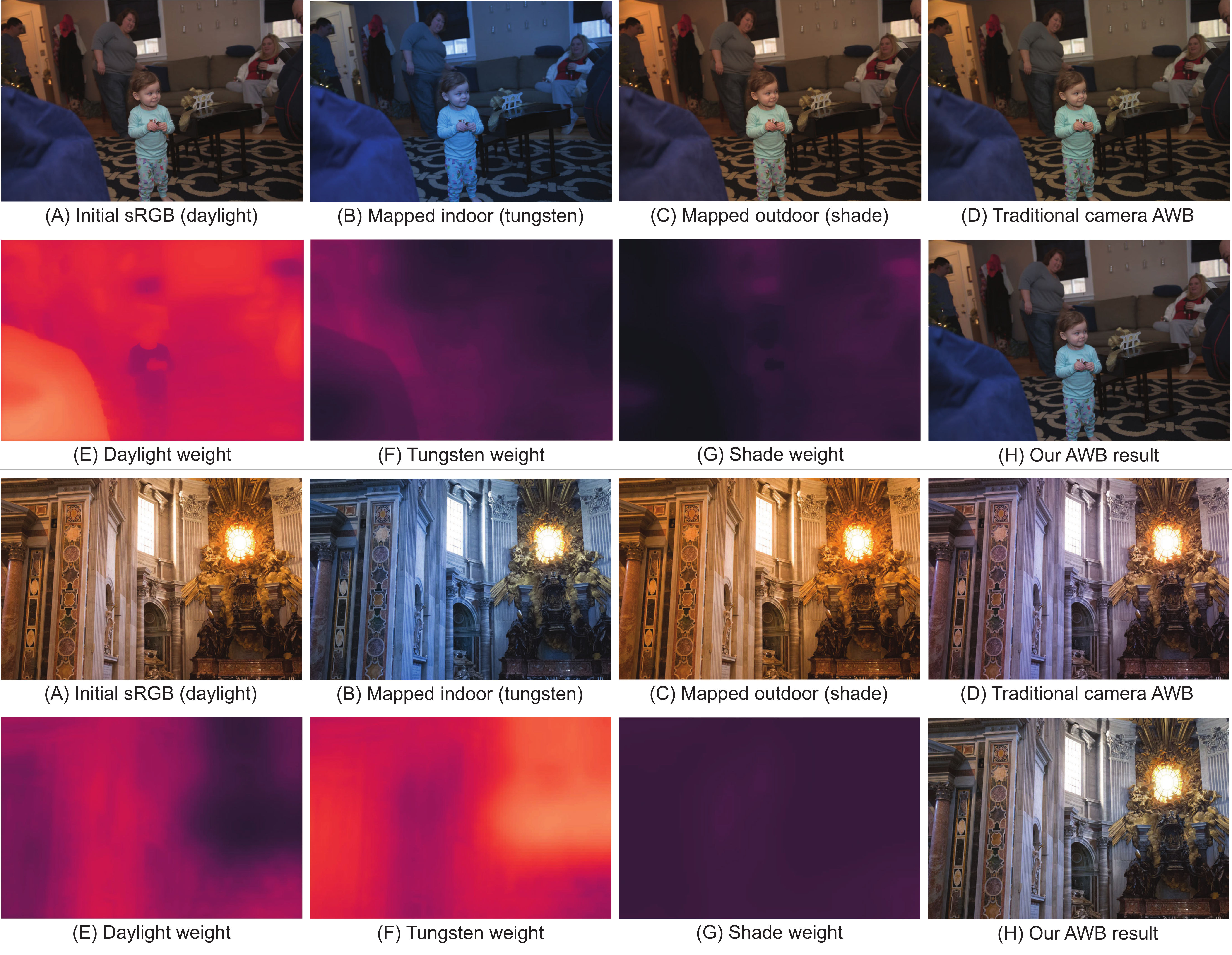}
\vspace{-6mm}
\caption{Predicted weighting maps. (A) Initial rendered image with daylight WB applied. (B-C) High-resolution images with indoor and outdoor WB settings after color mapping. (D) Traditional camera AWB. (E-G) predicted weights for each WB setting. (H) Our final AWB. Images are from the MIT-Adobe 5K dataset \cite{fivek}. \label{fig:weights}}
\end{figure*}

%% ABLATION FIGURES
\begin{figure}[t]
\centering
\includegraphics[width=\linewidth]{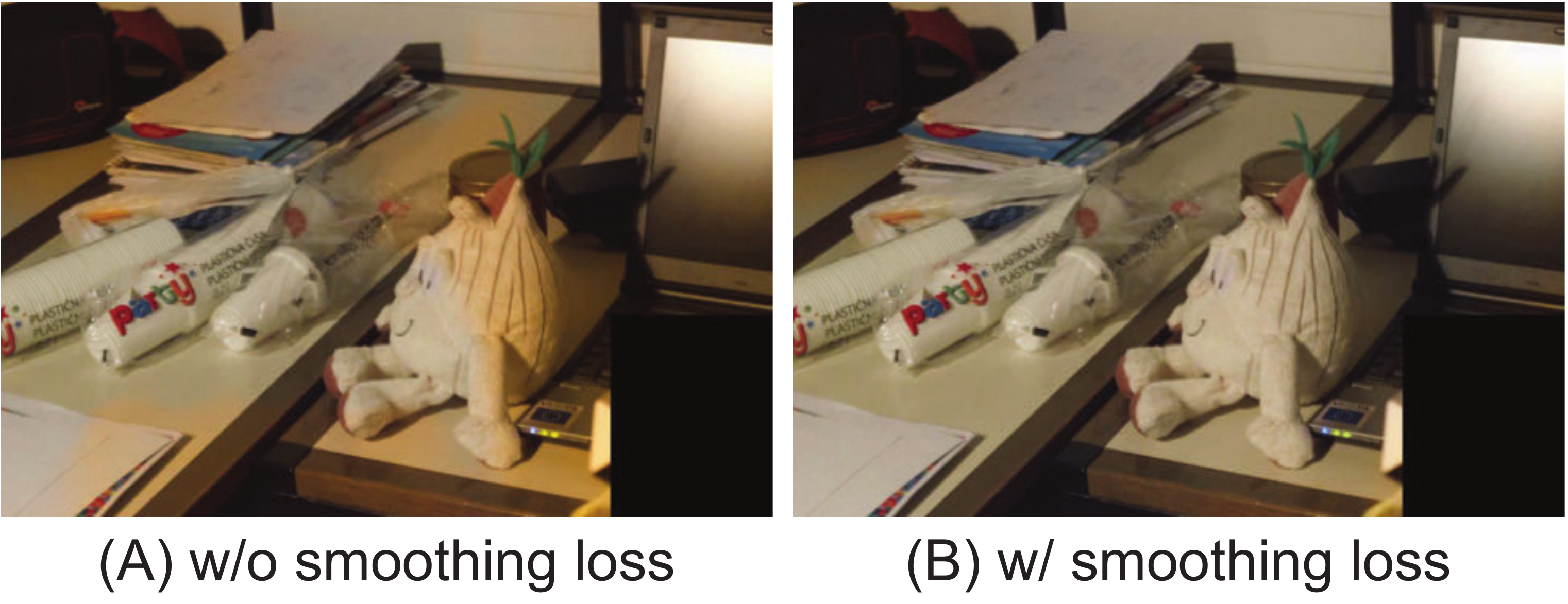}
\vspace{-6mm}
\caption{Impact of the smoothing loss term (\smoothingloss). (A) Result without \smoothingloss. (B) Results with \smoothingloss. Input image are from the Cube+ dataset \cite{banic2017unsupervised}.
\label{fig:smoothing}}
\vspace{-4mm}
\end{figure}

\subsection{Evaluation Sets} \label{subsec:evaluation-sets}

%% Qualitative FIGURES
\begin{figure*}[t]
\centering
\includegraphics[width=\linewidth]{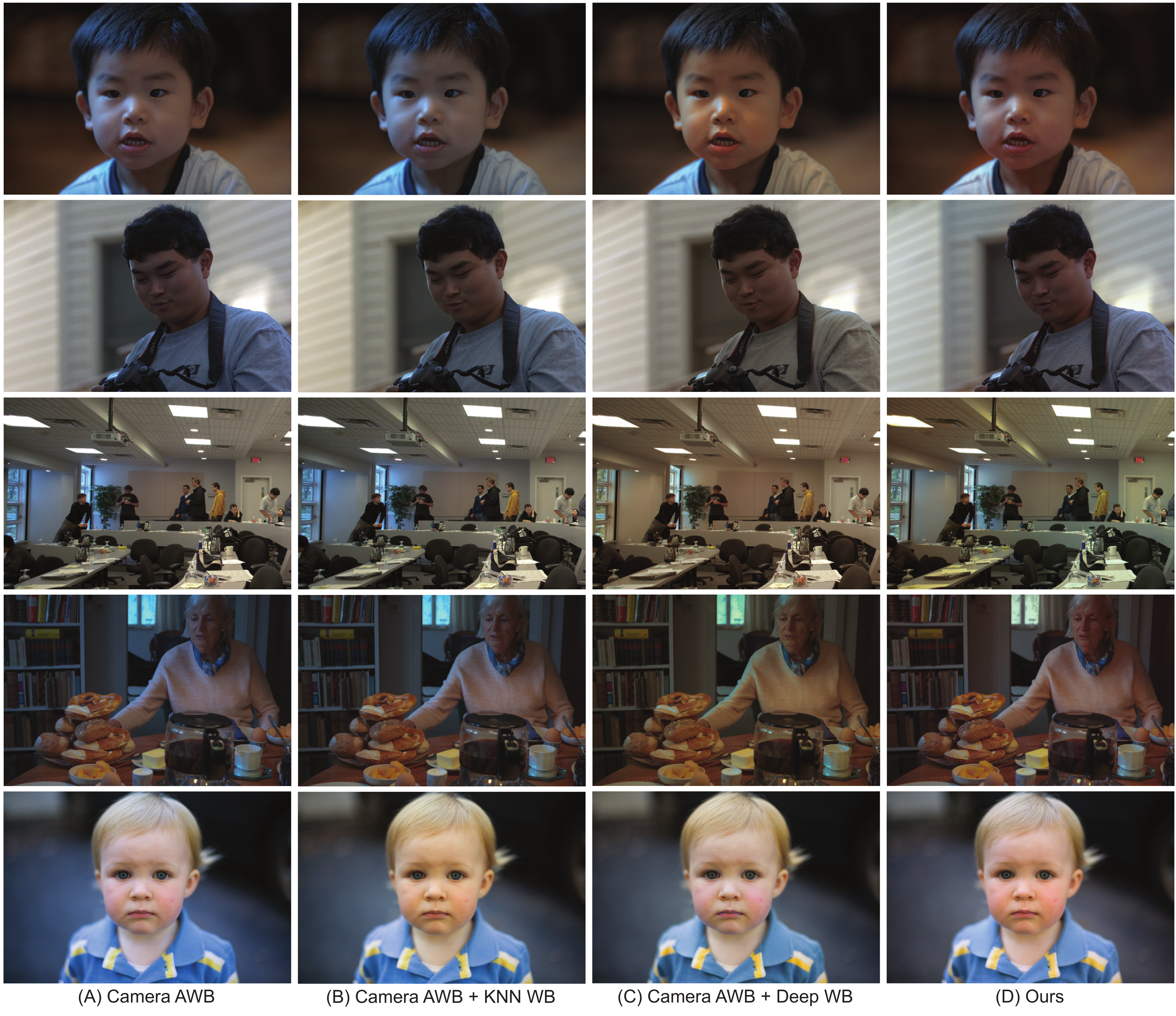}
\vspace{-6mm}
\caption{Qualitative comparisons with other AWB methods on the MIT-Adobe 5K dataset \cite{fivek}. Shown are the results of the following methods: gray pixel \cite{GP}, grayness index \cite{GI}, interactive WB \cite{afifi2020interactive}, KNN WB \cite{afifi2019color}, deep WB \cite{afifi2020deep}, and our method. \label{fig:results-fivek}}
\end{figure*}

We used three different datasets to evaluate our method. The scenes and camera models present in these evaluation sets were not part of our training data.

\paragraph{Cube+ dataset \cite{banic2017unsupervised}}
As the Cube+ dataset \cite{banic2017unsupervised} is commonly used by prior color constancy work, we used the linear raw images provided in the Cube+ dataset to evaluate our AWB method for single-illuminant scenes. Specifically, we used the raw images in order to generate the small images to feed them into our DNN. 

\begin{figure}[t]
\centering
\includegraphics[width=\linewidth]{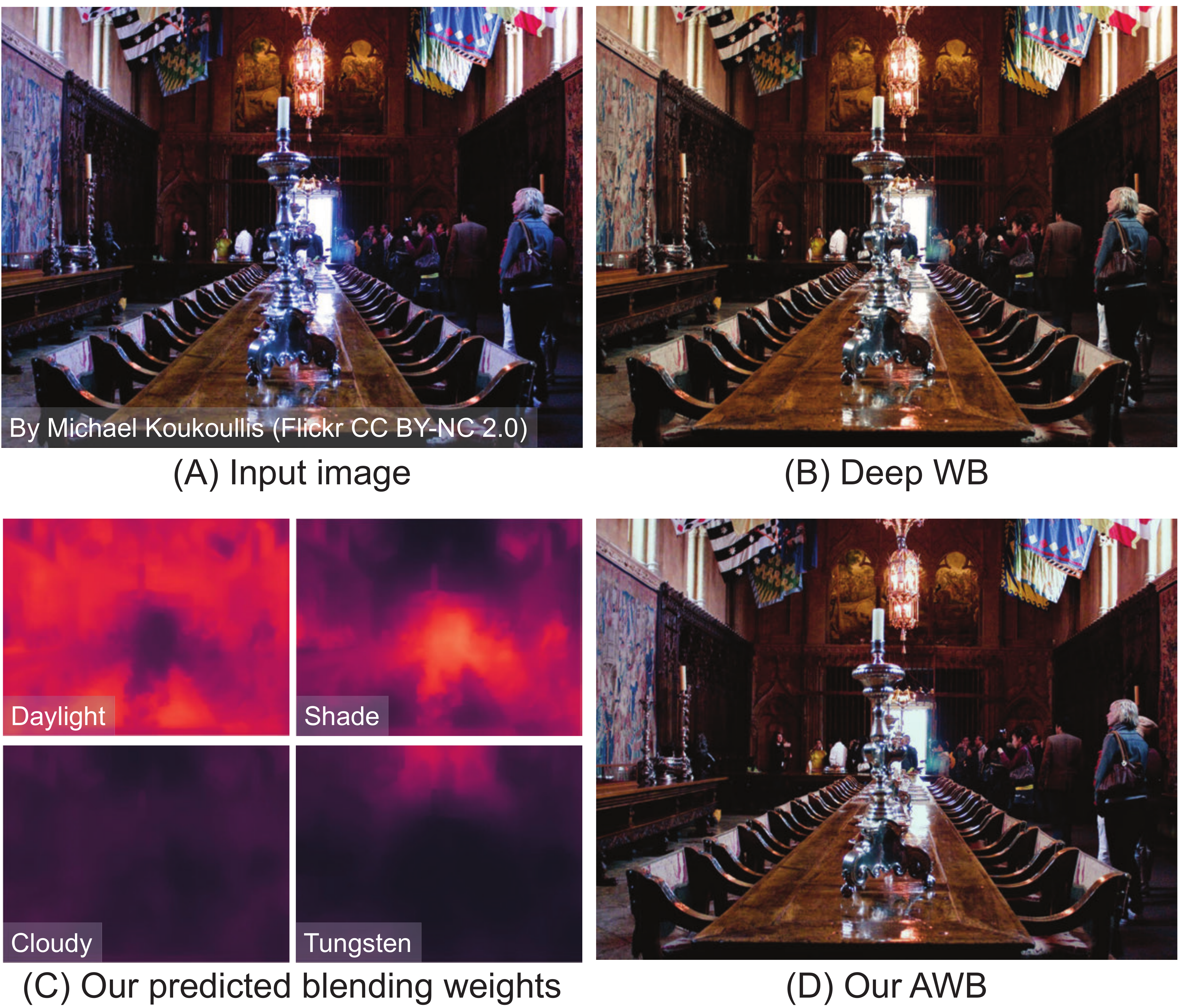}
\vspace{-6mm}
\caption{Post-capture WB correction. By employing an off-the-shelf method for global WB editing (e.g., \cite{afifi2020deep}), we can employ our method to correct mixed-illuminant scene colors in post-capture time. Input image is from Flickr.\label{fig:results-flickr}}\vspace{-6mm}
\end{figure}

\paragraph{MIT-Adobe 5K dataset \cite{fivek}}
We selected a set of mixed-illuminant scenes from the MIT-Adobe 5K dataset \cite{fivek} for qualitative comparisons. Similar to the Cube+ dataset, the MIT-Adobe 5K dataset \cite{fivek} provides the linear raw DNG file of each image, which facilitates our small image rendering process. This set is used to compare our results against the traditional camera AWB results by using the ``as-shot'' WB values stored in each DNG raw file.

\paragraph{Our mixed-illuminant test set}

As a part of our contribution, we generated a synthetic testing set to quantitatively evaluate mixed-illuminant scene WB methods. To generate our test set, we started with 3D scenes modeled in Autodesk 3Ds Max \cite{3dsmax}. We added multiple light sources in each scene, such that each scene has at least two types of light sources (e.g., indoor and outdoor lighting). For each scene, we set the virtual camera in different poses and used different color temperatures (e.g., 28500 K, 5500 K) for the camera AWB setting. Then, we rendered the final scene images using Vray rendering \cite{vray} to generate realistic photographs. In total, we rendered 150 images with mixed lighting conditions. Each image comes with its associated ground-truth image. To render the ground-truth images, we set the color temperature of our virtual camera's WB and all light sources in the scene to 5500 K (i.e., daylight). Note that setting the color temperature does not affect the intensity of the final rendered image pixels, but only changes the lighting colors; see Figure \ref{fig:dataset}. In comparison with existing multi-illuminant datasets (e.g., \cite{beigpour2013multi, beigpour2016multi, murmann2019dataset, hao2019evaluating, hao2020multi}), our test set is the first set that includes images rendered with different WB settings along with the corresponding ground-truth images rendered with correctly white-balanced pixels in the rendering pipeline. See the appendix for further discussion. 

\subsection{Results} \label{subsec:results}

Figure \ref{fig:weights} shows examples of the generated blending weights. In this figure, we compare our results, shown in Figure \ref{fig:weights}-(H), with the traditional camera AWB results, shown in Figure \ref{fig:weights}-(D). Figure \ref{fig:weights} also shows the initial sRGB image rendered with the fixed WB setting (i.e., daylight), along with the mapped images to the predefined WB settings, as described in Equation \ref{eq:mapping}. As can be seen, our method learned to produce proper blending weights to locally white-balance each region in the captured scene. 

Table \ref{tab:results-cube} shows the results of our AWB method on  the single-illuminant Cube+ dataset \cite{banic2017unsupervised}. In this table, we also show the results of a set of ablation studies conducted to show the impact of some training options. Specifically, we show the results of training our network on the following patch sizes ($p$): $256\!\times\!256$, $128\!\times\!128$, and $64\!\times\!64$ pixels.

We also show the results of our method when training without using the smoothing loss term (\smoothingloss). The impact of \smoothingloss $ $ can also be shown in Figure \ref{fig:smoothing}. In addition, we show the results of training our method without keeping the same order of concatenated small images during training. That is, we shuffle the order of concatenated small images before feeding the images to the network during training to see if that would affect the accuracy. As shown in Table  \ref{tab:results-cube}, using a fixed order achieves better results as that helps our network to build knowledge on the correlation between pixel colors when rendered with different WB settings and the generated weighting maps. Table  \ref{tab:results-cube} also shows that when comparing our method with recent methods for WB correction, our method achieves very competitive results, while requiring small memory overhead compared to the state-of-the-art methods (e.g., \cite{afifi2020deep}). 

%% TABLES
% AWB
\begin{table*}[t]
\centering
\caption{Comparison on the single-illuminant Cube+ dataset \cite{banic2017unsupervised}. For our method, we report the results of several ablation studies, where we trained our model with different WB settings and patch sizes ($p$). The terms $\texttt{t}$, $\texttt{f}$, $\texttt{d}$, $\texttt{c}$, and $\texttt{s}$ refer to tungsten, fluorescent, daylight, cloudy, and shade, respectively. We followed prior work \cite{afifi2019color} and reported the mean, first, second (median), and third quantile (Q1, Q2, and Q3) of mean square error (MSE), mean angular error (MAE), and $\bigtriangleup$E 2000 \cite{sharma2005ciede2000}. For all diagonal-based methods, gamma linearization \cite{anderson1996proposal, ebner2007color} is applied. The top results are indicated with yellow and bold, while second and third best results are indicated with green and red, respectively. \label{tab:results-cube}}\vspace{-2mm}
\scalebox{0.76}{
\begin{tabular}{|l|c|c|c|c|c|c|c|c|c|c|c|c|c|}
\hline
\multicolumn{1}{|c|}{} & \multicolumn{4}{c|}{\textbf{MSE}} & \multicolumn{4}{c|}{\textbf{MAE}} & \multicolumn{4}{c|}{\textbf{$\boldsymbol{\bigtriangleup}$\textbf{E} 2000}} & \multicolumn{1}{|c|}{} \\ \cline{2-13}
\multicolumn{1}{|c|}{\multirow{-2}{*}{\textbf{Method}}} & \textbf{Mean} & \textbf{Q1} & \textbf{Q2} & \textbf{Q3} & \textbf{Mean} & \textbf{Q1} & \textbf{Q2} & \textbf{Q3} & \textbf{Mean} & \textbf{Q1} & \textbf{Q2} & \textbf{Q3} & \multicolumn{1}{|c|}{\multirow{-2}{*}{\textbf{Size}}} \\ \hline

FC4 \cite{hu2017fc}  & 371.9 & 79.15 & 213.41 & 467.33 & 6.49\textdegree & 3.34\textdegree & 5.59\textdegree & 8.59\textdegree & 10.38 & 6.6 & 9.76 & 13.26 &  5.89 MB \\ \hline

Quasi-U CC \cite{bianco2019quasi} & 292.18& 15.57 & 55.41 & 261.58 & 6.12\textdegree & 1.95\textdegree & 3.88\textdegree & 8.83\textdegree & 7.25 & 2.89 & 5.21 & 10.37 &  622 MB \\ \hline

KNN WB \cite{afifi2019color} &  194.98 &  27.43 &  57.08 &  118.21 &  4.12\textdegree &  1.96\textdegree & 3.17\textdegree & 5.04\textdegree &  5.68 &  3.22 &  4.61 & \second 6.70 & 21.8 MB
\\ \hline											

Interactive WB \cite{afifi2020interactive} & \second 159.88 & 21.94 & 54.76 & 125.02 & 4.64\textdegree & 2.12\textdegree & 3.64\textdegree & 5.98\textdegree & 6.2 & 3.28 & 5.17 & 7.45 & \first \textbf{38 KB} \\ \hline

Deep WB \cite{afifi2020deep} &  \first \textbf{80.46} & 15.43 & 33.88 & \first \textbf{74.42} &  \first \textbf{3.45\textdegree} & 1.87\textdegree & 2.82\textdegree &  \first \textbf{4.26\textdegree} & \first \textbf{4.59} &  2.68 & 3.81 & \first \textbf{5.53} &  16.7 MB \\ \hdashline
\multicolumn{14}{|c|}{\cellcolor[HTML]{D4EBF2}\textbf{Our results}} \\ \hline

$p=256$, WB=$\{\texttt{t},\texttt{f}, \texttt{d}, \texttt{c}, \texttt{s}\}$ &  235.07 & 54.42 & 83.34 & 197.46 & 6.74\textdegree & 4.12\textdegree & 5.31\textdegree & 8.11\textdegree & 8.07 & 5.22 & 7.09 &  10.04  & \third 5.10 MB \\ \hline

$p=128$, WB=$\{\texttt{t},\texttt{f}, \texttt{d}, \texttt{c}, \texttt{s}\}$  &  176.38 & 16.96 & 35.91 & 115.50 & 4.71\textdegree & 2.10\textdegree & 3.09\textdegree & 5.92\textdegree &  5.77 & 3.01 & 4.27 & 7.71  & \third 5.10 MB \\ \hline

$p=64$, WB=$\{\texttt{t},\texttt{f}, \texttt{d}, \texttt{c}, \texttt{s}\}$ & \third 161.80 & \second 9.01 & \first\textbf{19.33} & \second 90.81 & \second 4.05\textdegree & \second 1.40\textdegree & \first \textbf{2.12\textdegree} & \second 4.88\textdegree & \second 4.89 & \second 2.16 & \first \textbf{3.10} & \third 6.78  & \third 5.10 MB \\ \hline

$p=64$, WB=$\{\texttt{t},\texttt{f}, \texttt{d}, \texttt{c}, \texttt{s}\}$, w/o \smoothingloss& 189.18 & \third 11.10 & \third 23.66 & 112.40 & 4.59\textdegree & \third 1.57\textdegree & \third 2.41\textdegree & 5.76\textdegree & 5.48 & \third 2.38 & \third 3.50 & 7.80 & \third 5.10 MB \\ \hline

$p=64$, WB=$\{\texttt{t},\texttt{f}, \texttt{d}, \texttt{c}, \texttt{s}\}$, w/ shuff.& 197.21 & 24.48 & 55.77 & 149.95 & 5.36\textdegree & 2.60\textdegree & 3.90\textdegree & 6.74\textdegree & 6.66 & 3.79 & 5.45 & 8.65 & \third 5.10 MB \\ \hline

$p=64$, WB=$\{\texttt{t}, \texttt{d}, \texttt{s}\}$ & 168.38 & \first \textbf{8.97} & \second 19.87 & \third 105.22 & \third 4.20\textdegree & \first \textbf{1.39\textdegree} & \second 2.18\textdegree & \third 5.54\textdegree & \third 5.03 & \first \textbf{2.07} & \second 3.12 &  7.19 & \second 5.09 MB \\ \hline

\end{tabular}
}
\end{table*}

In Table \ref{tab:results-our-set}, we report the results of our method, along with the results of recently published methods for WB correction in sRGB space, on our synthetic test set. As shown, our method achieves the best results over most evaluation metrics. Table \ref{tab:results-our-set} shows the results when training our method using different predefined sets of WB settings. 

Figure \ref{fig:results-fivek} shows qualitative comparisons on images from the MIT-Adobe 5K dataset \cite{fivek}. We compare our results, shown in Figure \ref{fig:results-fivek}-(D), against traditional camera AWB (Figure \ref{fig:results-fivek}-[A]), after applying the KNN-WB correction \cite{afifi2019color} (Figure \ref{fig:results-fivek}-[B]), and after applying the deep WB correction \cite{afifi2020deep} (Figure \ref{fig:results-fivek}-[C]). It is clear that our method provides better per-pixel correction compared to other alternatives. 

Our method has the limitation of requiring a modification to an ISP to render the small images. To overcome this, one could employ one of the sRGB WB editing methods to synthesize rendering our small images with the target predefined WB set in post-capture time. In Figure \ref{fig:results-flickr}, we illustrate this idea by employing the deep WB \cite{afifi2020deep} to generate the small images of a given sRGB camera-rendered image taken from Flickr. As shown, our method produces a better result compared to the camera-rendered image (i.e., traditional camera AWB) and the deep WB result for post-capture WB correction. 

\begin{figure}[t]
\centering
\includegraphics[width=\linewidth]{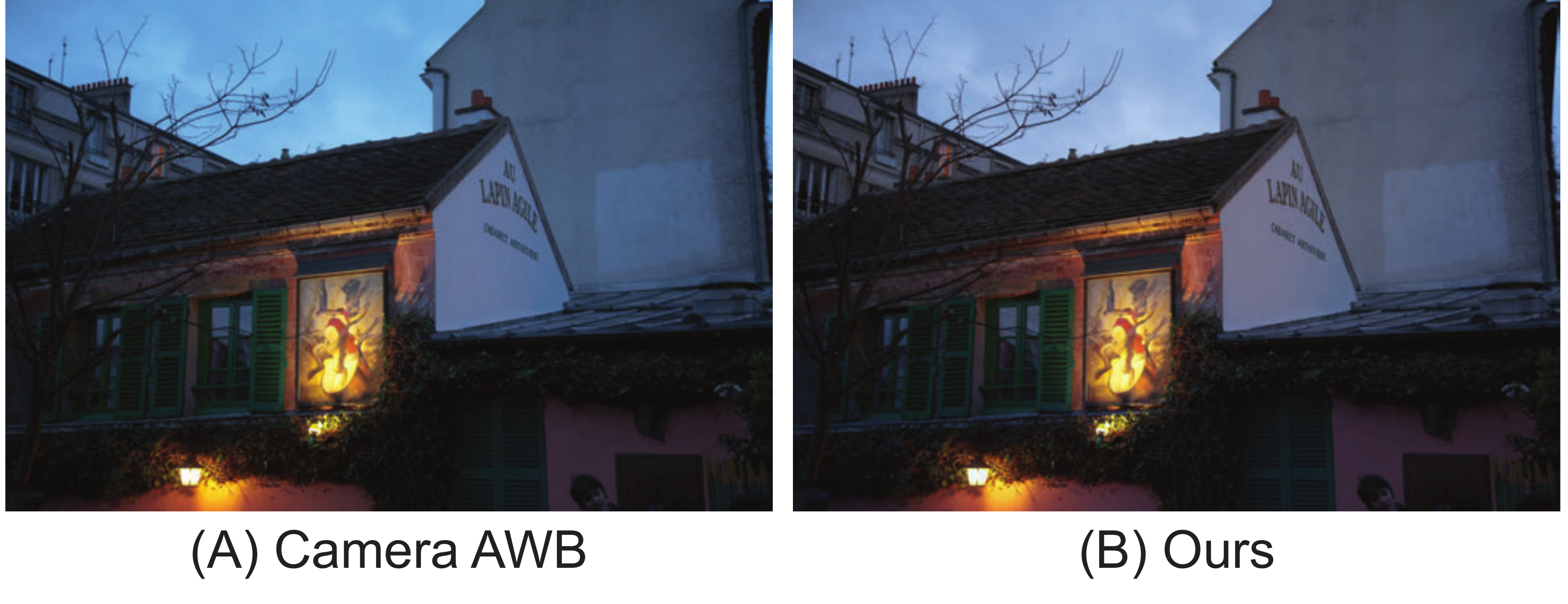}
\vspace{-6mm}
\caption{Failure examples. Our method fails in some cases, where it could not properly choose the correct WB setting for some regions in the input image. Input image is from the MIT-Adobe 5K dataset \cite{fivek}. \label{fig:failure-cases}}\vspace{-6mm}
\end{figure}

Lastly, Figure \ref{fig:failure-cases} shows a failure example of our method, where it could not properly correct colors of each pixel in the image. In such cases, our method produces results that are very similar to correcting only for a single illuminant, as it is bounded by our predefined set of WB settings.

% our dataset
\begin{table*}[t]
\centering
\caption{Comparison on our mixed-illuminant evaluation set. Highlights and symbols are the same as in Table \ref{tab:results-cube}. \label{tab:results-our-set}}\vspace{-2mm}
\scalebox{0.76}{
\begin{tabular}{|l|c|c|c|c|c|c|c|c|c|c|c|c|}
\hline
\multicolumn{1}{|c|}{} & \multicolumn{4}{c|}{\textbf{MSE}} & \multicolumn{4}{c|}{\textbf{MAE}} & \multicolumn{4}{c|}{\textbf{$\boldsymbol{\bigtriangleup}$\textbf{E} 2000}}  \\ \cline{2-13}
\multicolumn{1}{|c|}{\multirow{-2}{*}{\textbf{Method}}} & \textbf{Mean} & \textbf{Q1} & \textbf{Q2} & \textbf{Q3} & \textbf{Mean} & \textbf{Q1} & \textbf{Q2} & \textbf{Q3} & \textbf{Mean} & \textbf{Q1} & \textbf{Q2} & \textbf{Q3} \\ \hline

Gray pixel \cite{GP} & 4959.20 & 3252.14 & 4209.12 & 5858.69 & 19.67\textdegree & 11.92\textdegree & 17.21\textdegree & 27.05\textdegree & 25.13 & 19.07 & 22.62 & 27.46 \\ \hline

Grayness index \cite{GI} & 1345.47 & 727.90 & 1055.83 & 1494.81 & 6.39\textdegree & 4.72\textdegree & 5.65\textdegree & 7.06\textdegree & 12.84 & 9.57 & 12.49 & 14.60 \\ \hline

KNN WB \cite{afifi2019color} & 1226.57 & 680.65 & 1062.64 & 1573.89 & 5.81\textdegree & 4.29\textdegree & 5.76\textdegree & 6.85\textdegree & 12.00 & \third 9.37 & 11.56 & 13.61
\\ \hline											

Interactive WB \cite{afifi2020interactive} & 1059.88 & \second 616.24 &  896.90 & 1265.62 & 5.86\textdegree & 4.56\textdegree & 5.62\textdegree & 6.62\textdegree & \third 11.41 & \second 8.92 & \third 10.99 & 12.84 \\ \hline
 
Deep WB \cite{afifi2020deep} &  1130.598 & \third 621.00 & \third 886.32 & 1274.72 &  \first \textbf{4.53\textdegree} & \first \textbf{3.55\textdegree} & \second 4.19\textdegree & \first \textbf{5.21\textdegree} & \second 10.93 & \first \textbf{8.59} & \first \textbf{9.82} & \second 11.96 \\ \hdashline

Ours ($p=64$,  WB=$\{\texttt{t}, \texttt{d}, \texttt{s}\}$) & \first \textbf{819.47} & 655.88 & \first \textbf{845.79} & \second 1000.82 & 5.43\textdegree & 4.27\textdegree & 4.89\textdegree & 6.23\textdegree & \first \textbf{10.61} & 9.42 & \second 10.72 & \first \textbf{11.81} \\ \hline

Ours ($p=64$,  WB=$\{\texttt{t}, \texttt{f}, \texttt{d}, \texttt{c}, \texttt{s}\}$) & \third 938.02 & 757.49 & 961.55 &  \third 1161.52 & \second 4.67\textdegree & \second 3.71\textdegree & \first \textbf{4.14\textdegree} & \second 5.35\textdegree & 12.26 & 10.80 & 11.58 & \third 12.76 \\ \hline

Ours ($p=128$,  WB=$\{\texttt{t}, \texttt{d}, \texttt{s}\}$) & \second 830.20 & \first \textbf{584.77} & \second 853.01 & \first \textbf{992.56} & \third 5.03\textdegree & \third 3.93\textdegree & \third 4.78\textdegree & \third 5.90\textdegree & \third 11.41 & 9.76 & 11.39 & 12.53 \\ \hline

Ours ($p=128$,  WB=$\{\texttt{t}, \texttt{f}, \texttt{d}, \texttt{c}, \texttt{s}\}$) & 1089.69 & 846.21 & 1125.59 & 1279.39 & 5.64\textdegree & 4.15\textdegree & 5.09\textdegree & 6.50\textdegree & 13.75 & 11.45 & 12.58 & 15.59 \\ \hline

\end{tabular}
}
\end{table*}

%% FAILURE CASE FIGURE

\section{Conclusion}
We have presented an AWB method for mixed-illuminant scenes. Our method achieves local WB correction by producing weighting maps that blend between the same input image rendered with different WB settings. Our proposed method learns to produce these local weighting maps through a DNN. Our method is fast and thus can be deployed on board a camera ISP hardware. As a part of our contribution, we have proposed a synthetic test set of mixed-illuminant scenes with pixel-wise ground truth. Compared with other alternatives, we showed that our method produces promising results through both qualitative and quantitative evaluations.

\section*{Acknowledgement}
This study was funded in part by the
Canada First Research Excellence Fund for the Vision: Science
to Applications (VISTA) programme and an NSERC Discovery
Grant. The authors would like to thank Mohammed Hossam for the help in generating our synthetic test set.

\appendix

\section{Network Architecture} \label{sec:net}
\begin{figure}[t]
\centering
\includegraphics[width=\linewidth]{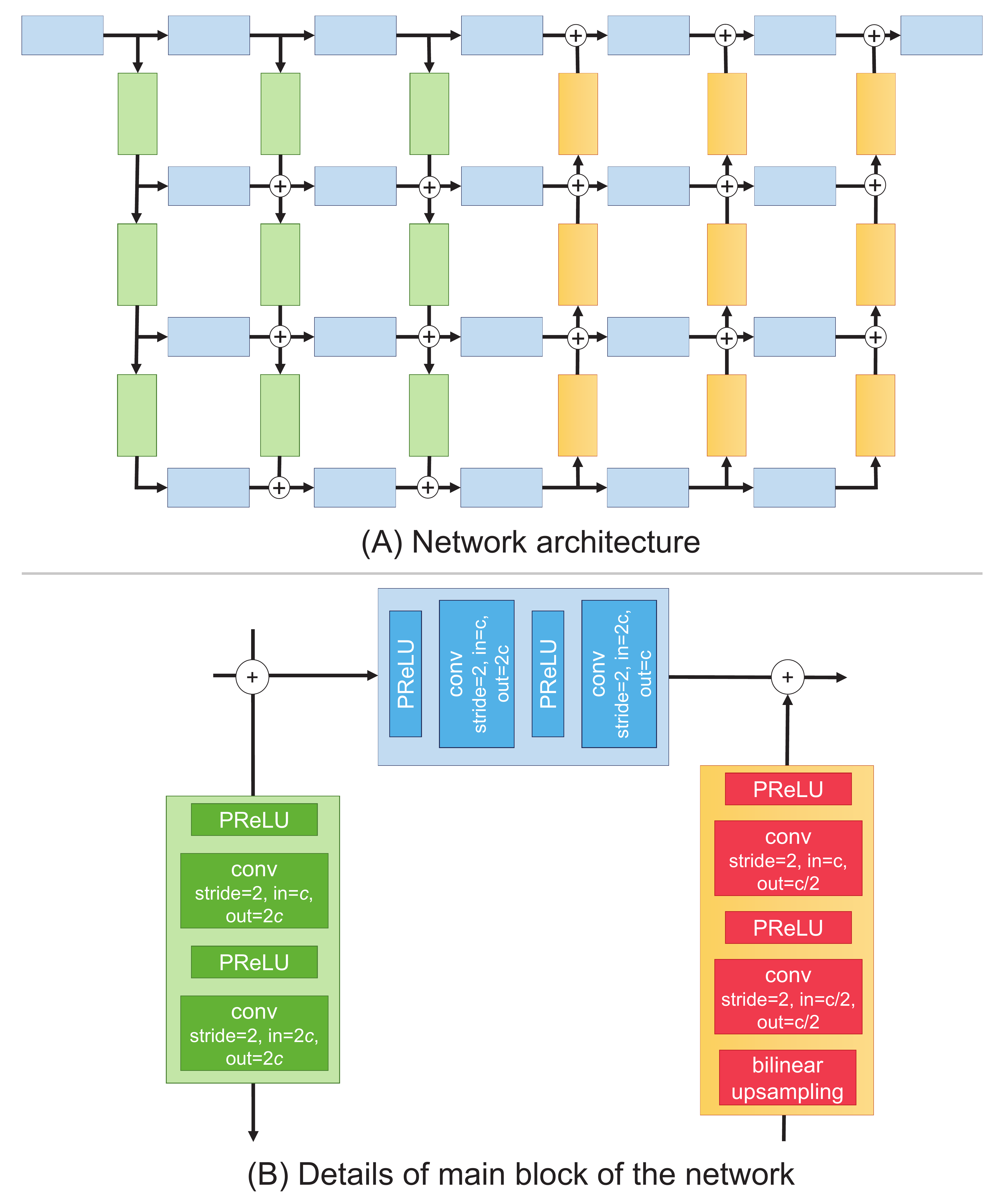}
\vspace{-6mm}
\caption{Network architecture. We adopted the GridNet architecture \cite{fourure2017residual, niklaus2018context}, as shown in (A). (B) shows the details of the residual, downsampling, and upsampling units. The symbol $c$ refer to the number of channels in each conv layer. \label{fig:net}}
\end{figure}

We adopted the GridNet architecture \cite{fourure2017residual, niklaus2018context}. Our network consists of six columns and four rows. As shown in Figure \ref{fig:net}, our network includes three main units, which are: the residual unit (shown in blue in Figure \ref{fig:net}), the downsampling unit (shown in green in Figure \ref{fig:net}), and the upsampling unit (shown in yellow in Figure \ref{fig:net}). 

The number of input/output channels, stride, and padding size of each conv layer are shown in Figure \ref{fig:net}-(B). The first residual unit of our network accepts concatenated input images with $3\!\times\!k$ channels, where $k$ refers to the number of images rendered with $k$ WB settings. For example, when using WB=$\{\texttt{t}, \texttt{d}, \texttt{s}\}$, the value of $k$ is three. Regardless of the value of $k$, we set the number of output channels of the first residual to unit to eight.

Each residual block (except for the first one), produces features with the same dimensions of the input feature. For the first three columns,  the dimensions of each feature received from the upper row are reduced by two, while the number of output channels is duplicated, as shown in the downsampling unit in Figure \ref{fig:net}-(B). In contrast, the upsampling unit (shown in Figure \ref{fig:net}-[B]) increases the dimensions of the received features by two in the last three columns. Lastly, the last residual unit produces output weights with $k$ channels.  

\section{Our Synthetic Test Set} \label{sec:test-set}

\begin{figure}[t]
\centering
\includegraphics[width=\linewidth]{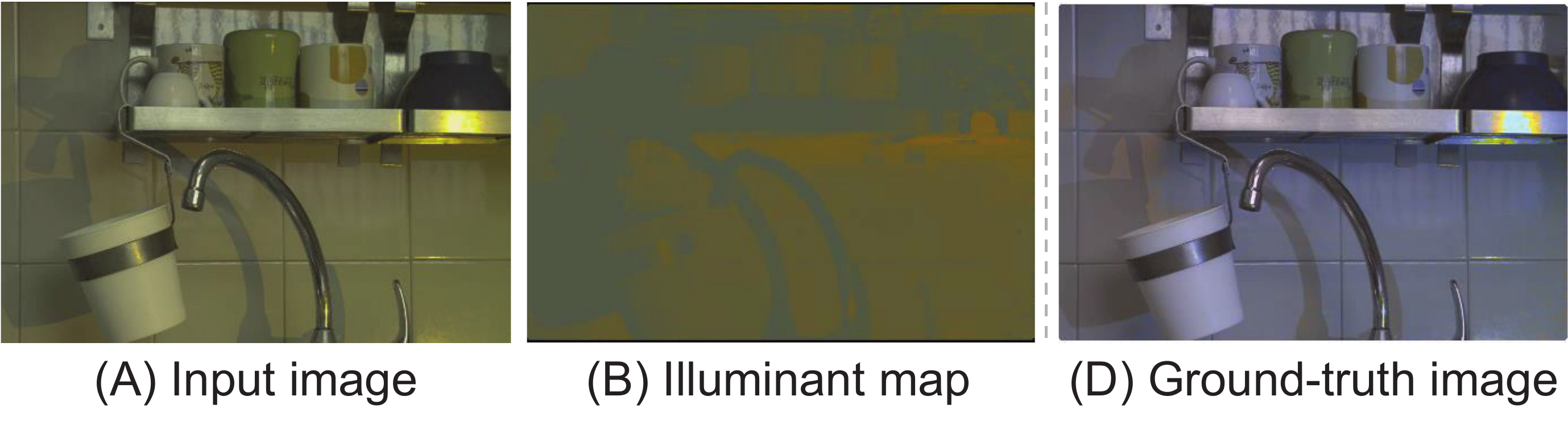}
\vspace{-6mm}
\caption{Example from the two-illuminant dataset \cite{beigpour2013multi}. (A) Input sRGB image. (B) Provided ground-truth illuminant map. (C) Generated ground-truth image using the illuminant map in (B).  \label{fig:dataset_comparison}}
\end{figure}

\begin{figure*}[t]
\centering
\includegraphics[width=\linewidth]{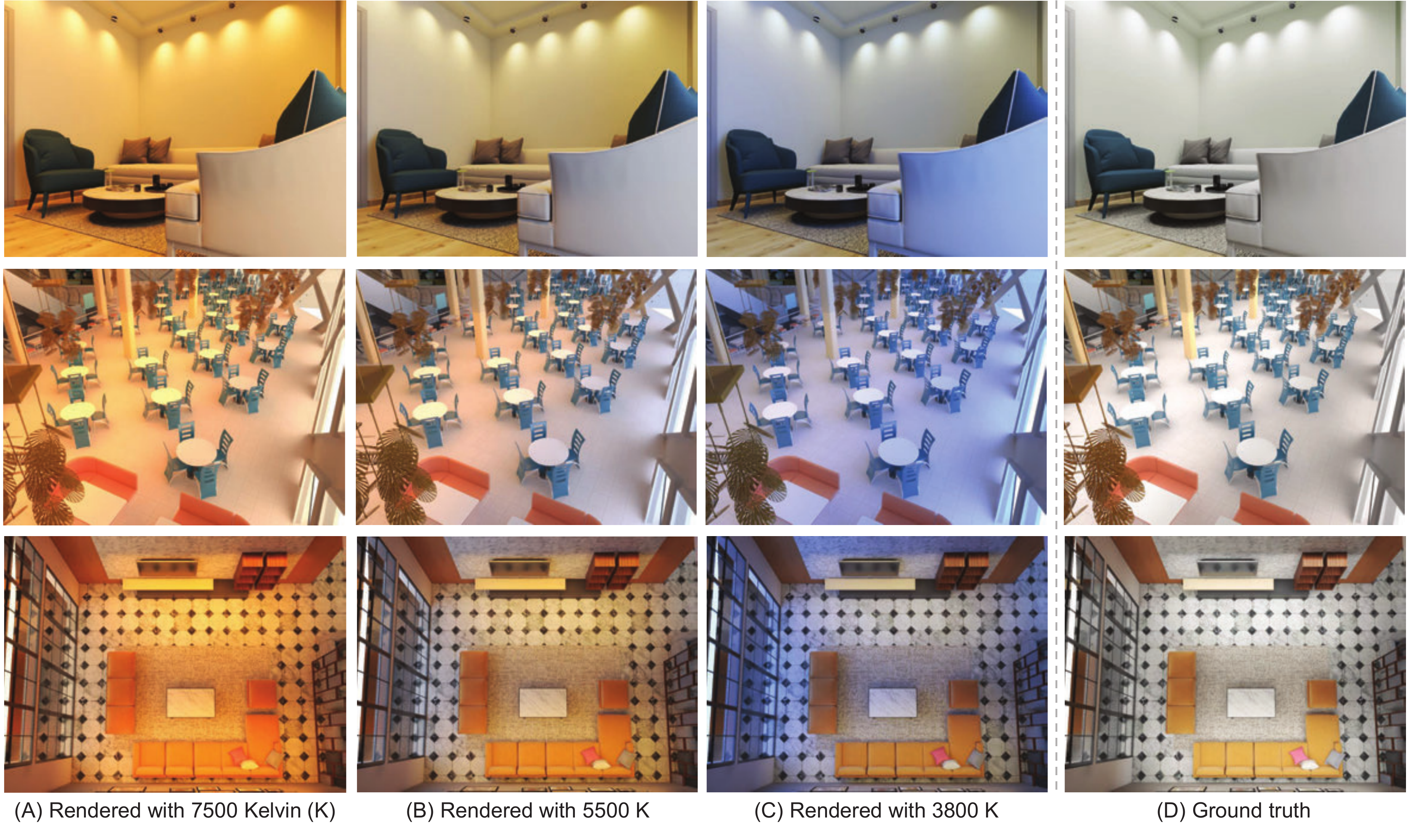}
\vspace{-6mm}
\caption{Examples from our synthetic test set. (A-C) Rendered images with different color temperatures that are associated to the following WB settings: shade, daylight, and fluorescent, respectively \cite{afifi2019else}. (D) Ground-truth image. \label{fig:dataset_supp}}
\end{figure*}

% ablation on post-processing
\begin{table*}[t]
\centering
\caption{Impact of ensembling and edge-aware smoothing (EAS) \cite{barron2016fast} at inference time. In this set of experiments, we used WB=$\{\texttt{t}, \texttt{d}, \texttt{s}\}$ with training patch-size $p=64$. We reported the mean, first, second (median), and third quantile (Q1, Q2, and Q3) of mean square error (MSE), mean angular error (MAE), and $\boldsymbol{\bigtriangleup}$\textbf{E} 2000 \cite{sharma2005ciede2000}. The top results are indicated with yellow and bold. \label{tab:ablation}}\vspace{-2mm}
\scalebox{0.79}{
\begin{tabular}{|l|c|c|c|c|c|c|c|c|c|c|c|c|}
\hline
\multicolumn{1}{|c|}{} & \multicolumn{4}{c|}{\textbf{MSE}} & \multicolumn{4}{c|}{\textbf{MAE}} & \multicolumn{4}{c|}{\textbf{$\boldsymbol{\bigtriangleup}$\textbf{E} 2000}}  \\ \cline{2-13}
\multicolumn{1}{|c|}{\multirow{-2}{*}{\textbf{Method}}} & \textbf{Mean} & \textbf{Q1} & \textbf{Q2} & \textbf{Q3} & \textbf{Mean} & \textbf{Q1} & \textbf{Q2} & \textbf{Q3} & \textbf{Mean} & \textbf{Q1} & \textbf{Q2} & \textbf{Q3} \\ \hline

w/o ensembling, w/o EAS & 849.33 & 694.68 & 846.80 & 1051.24 & 5.58\textdegree & 4.55\textdegree & 5.05\textdegree & 6.46\textdegree & 10.73 & 9.51 & 10.76 & 11.95 \\ \hline

w/ ensembling, w/o EAS & 831.41 & 662.25 & 855.24 & 1005.04 & \first \textbf{5.42\textdegree} & 4.38\textdegree & 4.98\textdegree & \first \textbf{6.08\textdegree} & 10.64 & 9.46 & \first \textbf{10.66} & 11.91 \\ \hline

w/ ensembling, w/ EAS & \first \textbf{819.47} & \first \textbf{655.88} & \first \textbf{845.79} &\first \textbf{1000.82} & 5.43\textdegree & \first \textbf{4.27\textdegree} & \first \textbf{4.89\textdegree} & 6.23\textdegree & \first \textbf{10.61} & \first \textbf{9.42} & 10.72 & \first \textbf{11.81}  \\ \hline

\end{tabular}
}
\end{table*}

As mentioned in the main paper, we have generated a set of 150 images with mixed illuminations. The ground-truth of each image is provided by rendering the same scene with a fixed color temperature used for all light sources in the scene and the camera AWB. 

Existing paired multi-illuminant datasets provide ground-truth images as either color maps or albedo layers (i.e., reflectance) \cite{bleier2011color, beigpour2013multi, hao2019evaluating, hao2020multi}. For instance, the two-illuminant dataset \cite{beigpour2013multi} includes 78 images (58 laboratory images taken  under close-to-ideal conditions and 20 real-world images). Each test image has a corresponding ground-truth illuminant color map, as shown in Figure \ref{fig:dataset_comparison}-(B). Unfortunately, the two-illuminant dataset \cite{beigpour2013multi} cannot be used by our method as it does not provide the same scene rendered with different WB settings. The provided raw images are in the PNG format and there are no DNG metadata provided to render such images to sRGB with different WB settings. In addition, creating the final ground-truth image in sRGB space, given the ground-truth illuminant map, does not always give satisfying results; see the specular region shown in Figure \ref{fig:dataset_comparison}-(C).

Recently, the MIST dataset \cite{hao2019evaluating, hao2020multi} was generated by rendering 3D scenes using Blender Cycles \cite{blender}. The MIST dataset was proposed to handle the limitations of existing multi-illuminant datasets (e.g., \cite{beigpour2013multi, gijsenij2011color, beigpour2016multi}) by providing accurate ground-truth image intrinsic properties (i.e., albedo, diffuse, and specular layers)  after an accurate measurement of the illumination at every point in each 3D scene. Despite its useful impact on testing image intrinsic decomposition methods, there is no ground-truth image provided for our task---namely, pixel-wise image white balancing.

Due to the aforementioned limitations of existing multi-illuminant test sets, we generated our test set with accurate ground-truth images for evaluating WB methods targeting mixed-illuminant scenes; see Figure \ref{fig:dataset_supp}

\section{Additional Ablation Studies} \label{sec:additional-ablations}

\begin{figure*}[t]
\centering
\includegraphics[width=\linewidth]{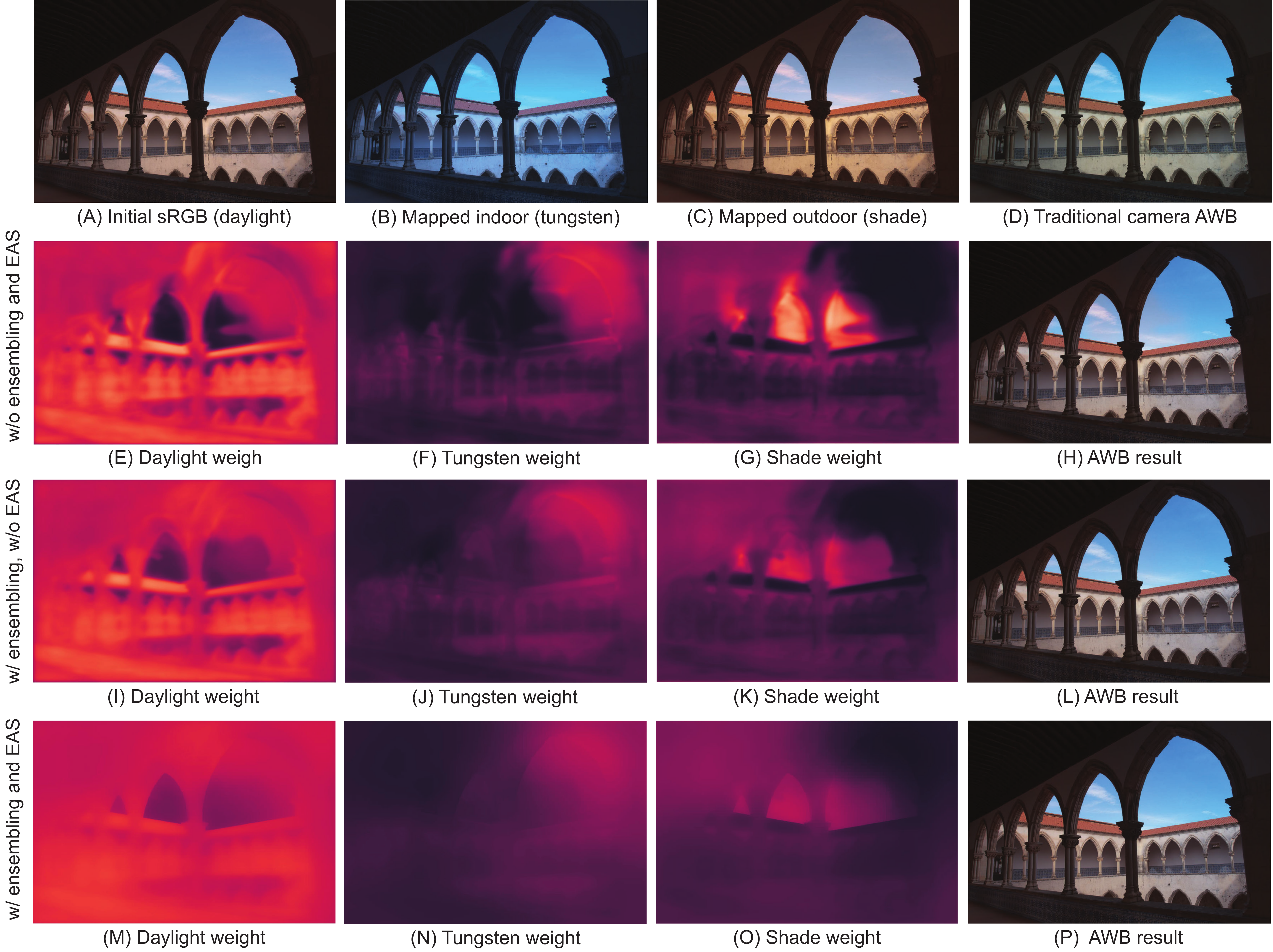}
\vspace{-6mm}
\caption{Qualitative examples showing the impact of ensembling and edge-aware smoothing (EAS) at inference time. The first row shows:  (A) the initial rendered image with daylight white-balance setting, (B-C) indoor and outdoor high-resolution images after mapping, and  (D) the result of traditional camera AWB correction. The second row shows: (E-G) the predicted weights without ensembling nor the EAS post-processing, along with the final AWB result after blending in (H). The third row shows the results when using the ensemble processing. The fourth row shows the results when using ensembling and EAS post-processing. Input images are from the the MIT-Adobe 5K dataset \cite{fivek}. \label{fig:ablation_1}}
\end{figure*}

\begin{figure*}[t]
\centering
\includegraphics[width=\linewidth]{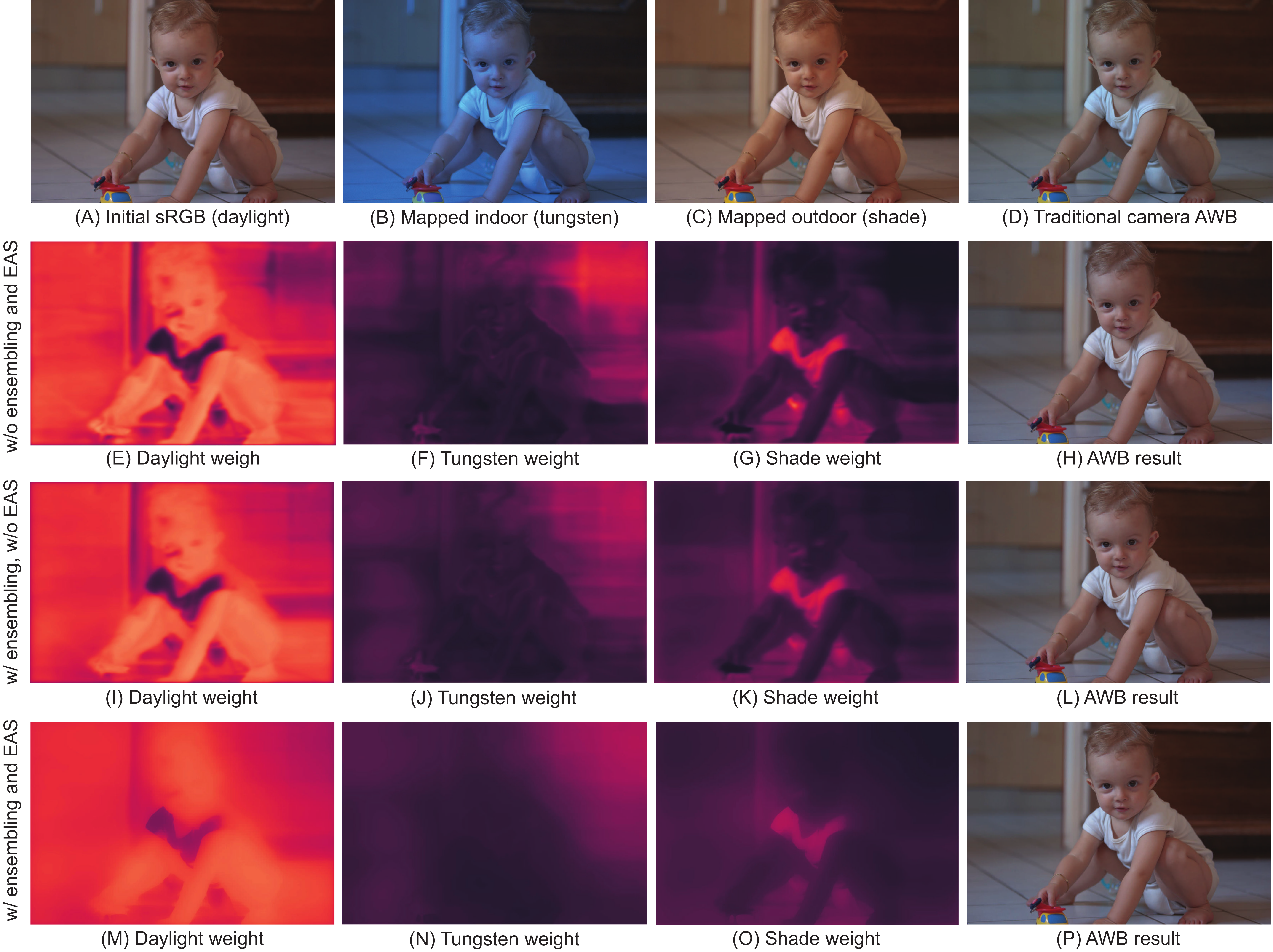}
\vspace{-6mm}
\caption{Additional qualitative examples showing the impact of ensembling and edge-aware smoothing (EAS) at inference time. The first row shows:  (A) the initial rendered image with daylight white-balance setting, (B-C) indoor and outdoor high-resolution images after mapping, and  (D) the result of traditional camera AWB correction. The second row shows: (E-G) the predicted weights without ensembling nor the EAS post-processing, along with the final AWB result after blending in (H). The third row shows the results when using the ensemble processing. The fourth row shows the results when using ensembling and EAS post-processing. Input images are from the the MIT-Adobe 5K dataset \cite{fivek}. \label{fig:ablation_2}}
\end{figure*}

In the main paper, we showed the results of ablation studies on the impact of different settings, including the size of training patches ($p$), the WB settings used to render input small images, and the smoothing loss term (\smoothingloss).

In this section, we show additional ablation study conducted to show the impact of the ensembling step and the post-processing edge-aware smoothing (EAS) step \cite{barron2016fast}. Table \ref{tab:ablation} shows the results of our method with and without the ensemble approach and the EAS post-processing step. The size of the input images is $384\!\times\!384$ pixels when ensembling is applied; otherwise, we used images of $256\!\times\!256$ pixels. Empirically, we found that image size of $256\!\times\!256$ pixels or $128\!\times\!128$ pixels give always the best results when the ensemble approach is not used. 

Figures \ref{fig:ablation_1} and \ref{fig:ablation_2}  show qualitative comparisons of our results with and without the ensembling and the EAS steps. As shown, when the ensemble testing is used, our predicted weights have more local coherence, which is further improved when using the EAS step.

\section{Additional Results}\label{sec:additional-results}
\begin{figure*}[t]
\centering
\includegraphics[width=\linewidth]{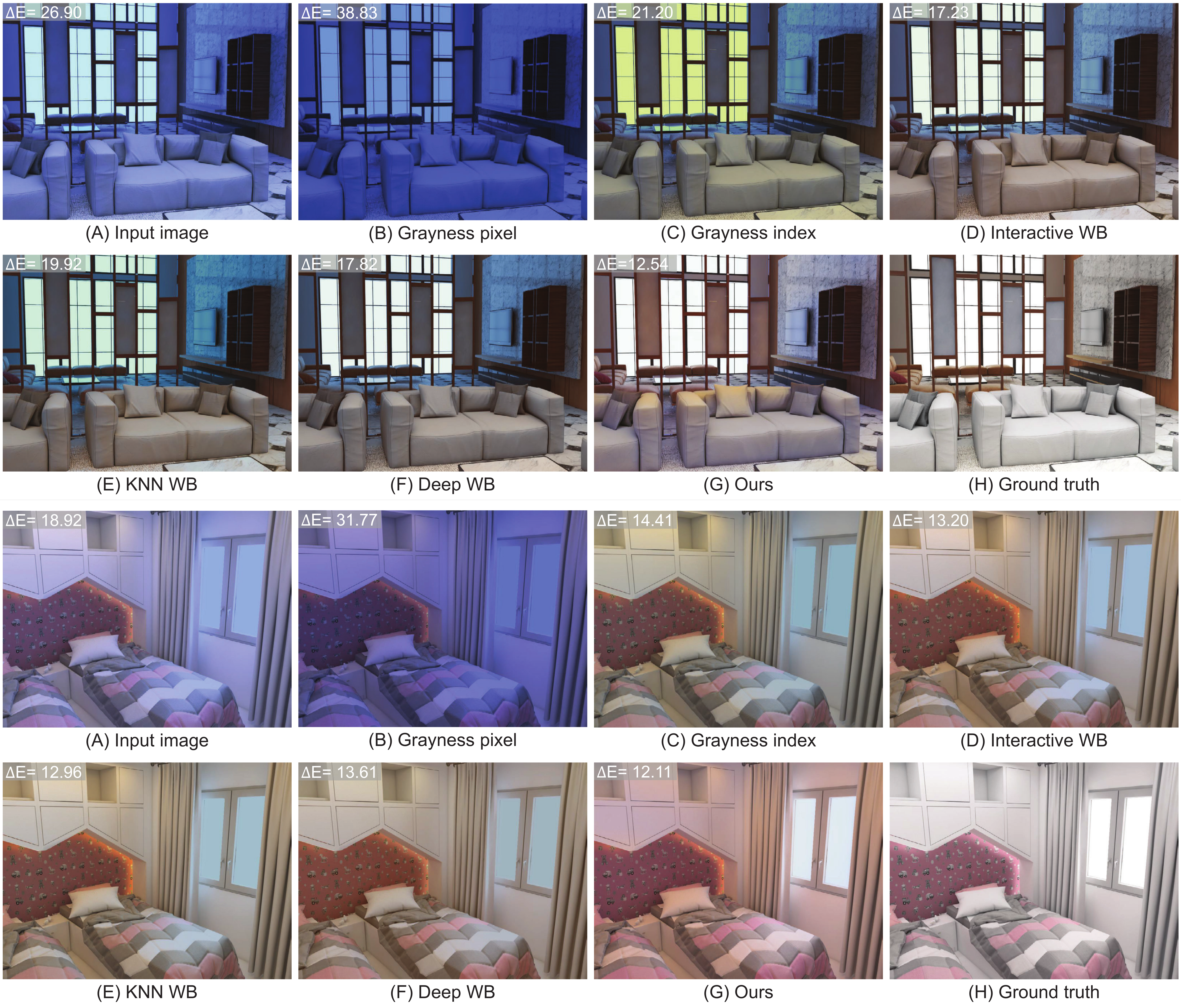}
\vspace{-6mm}
\caption{Qualitative comparisons with other AWB methods on our mixed-illuminant evaluation set. Shown are the results of the following methods: gray pixel \cite{GP}, grayness index \cite{GI}, interactive WB \cite{afifi2020interactive}, KNN WB \cite{afifi2019color}, deep WB \cite{afifi2020deep}, and our method.  \label{fig:results-our-set-supp}}
\end{figure*}

In the main paper, we reported quantitative results on our test set of our method and other methods for WB correction, which are: gray pixel \cite{GP}, grayness index \cite{GI}, KNN WB \cite{afifi2019color}, Interactive WB \cite{afifi2020interactive}, and Deep WB \cite{afifi2020deep}. Here, we show qualitative comparisons between our method and the aforementioned methods in Figure \ref{fig:results-our-set-supp}.

Finally, we show additional results on the MIT-Adobe 5K dataset \cite{fivek} in Figure \label{fig:results-fivek-supp}. Note that none of cameras/images in these sets were used in training either our method or the other methods.

\begin{figure*}[t]
\centering
\includegraphics[width=\linewidth]{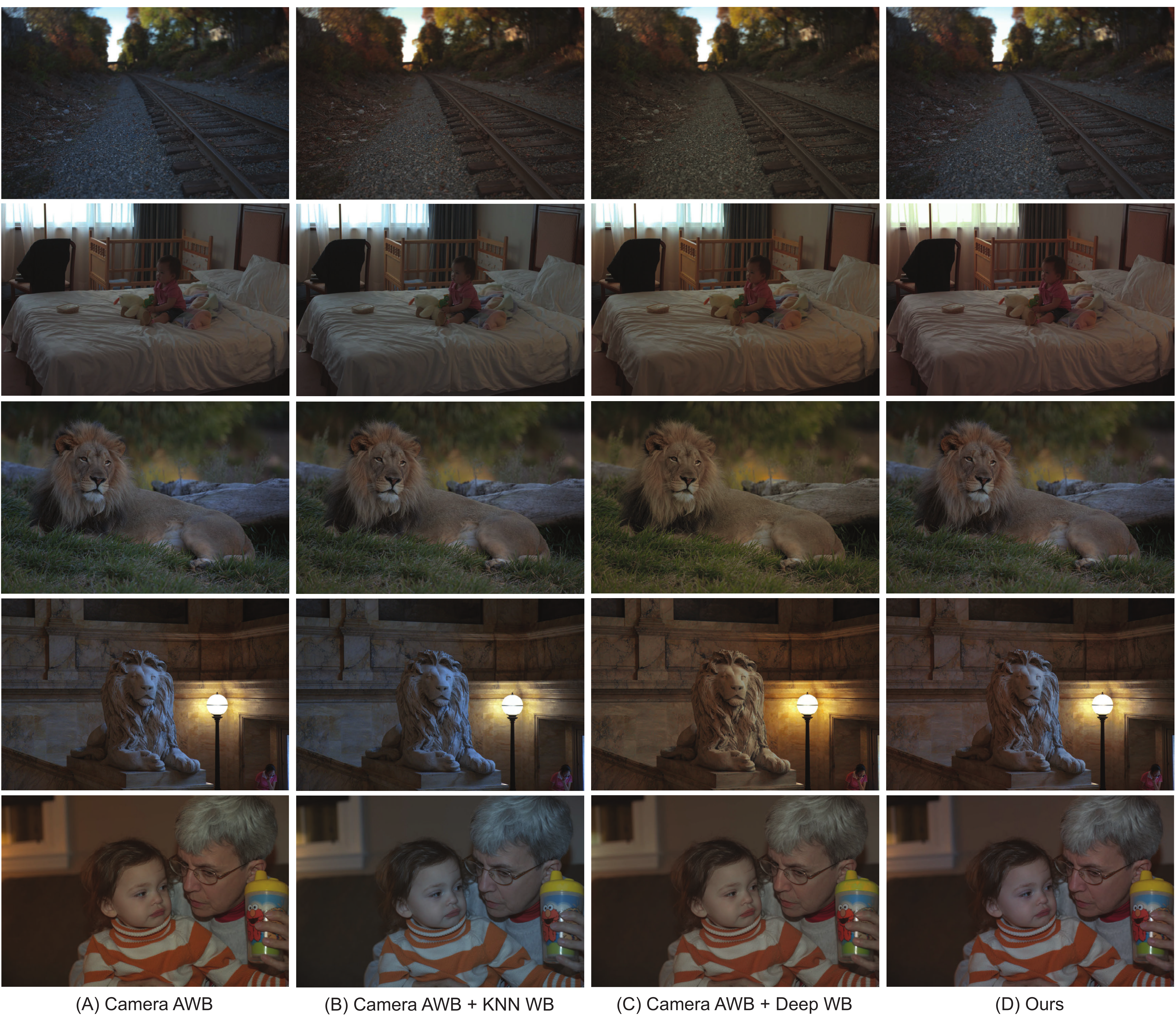}
\vspace{-6mm}
\caption{Additional qualitative comparisons with other AWB methods on the MIT-Adobe 5K dataset \cite{fivek}. Shown are the results of the following methods: gray pixel \cite{GP}, grayness index \cite{GI}, interactive WB \cite{afifi2020interactive}, KNN WB \cite{afifi2019color}, deep WB \cite{afifi2020deep}, and our method.  \label{fig:results-fivek-supp}}
\end{figure*}

{\small
\bibliographystyle{ieee_fullname}
\bibliography{ref}

\begin{thebibliography}{10}\itemsep=-1pt

\bibitem{blender}
Blender.
\newblock \url{https://www.blender.org}.
\newblock Accessed: 2021-07-20.

\bibitem{vray}
V-ray for 3ds max.
\newblock \url{https://www.chaosgroup.com/3d-rendering-software}.
\newblock Accessed: 2021-07-20.

\bibitem{c5}
Mahmoud Afifi, Jonathan~T Barron, Chloe LeGendre, Yun-Ta Tsai, and Francois
  Bleibel.
\newblock Cross-camera convolutional color constancy.
\newblock In {\em ICCV}, 2021.

\bibitem{afifi2019else}
Mahmoud Afifi and Michael~S Brown.
\newblock What else can fool deep learning? addressing color constancy errors
  on deep neural network performance.
\newblock In {\em ICCV}, 2019.

\bibitem{afifi2020deep}
Mahmoud Afifi and Michael~S Brown.
\newblock Deep white-balance editing.
\newblock In {\em CVPR}, 2020.

\bibitem{afifi2020interactive}
Mahmoud Afifi and Michael~S Brown.
\newblock Interactive white balancing for camera-rendered images.
\newblock In {\em Color and Imaging Conference}, 2020.

\bibitem{afifi2019color}
Mahmoud Afifi, Brian Price, Scott Cohen, and Michael~S Brown.
\newblock When color constancy goes wrong: Correcting improperly white-balanced
  images.
\newblock In {\em CVPR}, 2019.

\bibitem{afifi2019tuning}
Mahmoud Afifi, Abhijith Punnappurath, Abdelrahman Abdelhamed, Hakki~Can
  Karaimer, Abdullah Abuolaim, and Michael~S Brown.
\newblock Color temperature tuning: Allowing accurate post-capture
  white-balance editing.
\newblock In {\em Color and Imaging Conference}, 2019.

\bibitem{afifi2019projective}
Mahmoud Afifi, Abhijith Punnappurath, Graham Finlayson, and Michael~S Brown.
\newblock As-projective-as-possible bias correction for illumination estimation
  algorithms.
\newblock {\em JOSA A}, 36(1):71--78, 2019.

\bibitem{anderson1996proposal}
Matthew Anderson, Ricardo Motta, Srinivasan Chandrasekar, and Michael Stokes.
\newblock Proposal for a standard default color space for the {I}nternet \--
  s{R}{G}{B}.
\newblock In {\em Color and Imaging Conference}, 1996.

\bibitem{banic2017unsupervised}
Nikola Bani{\'c} and Sven Lon{\v{c}}ari{\'c}.
\newblock Unsupervised learning for color constancy.
\newblock {\em arXiv preprint arXiv:1712.00436}, 2017.

\bibitem{ccc}
Jonathan~T Barron.
\newblock Convolutional color constancy.
\newblock In {\em ICCV}, 2015.

\bibitem{barron2016fast}
Jonathan~T Barron and Ben Poole.
\newblock The fast bilateral solver.
\newblock In {\em ECCV}, 2016.

\bibitem{ffcc}
Jonathan~T Barron and Yun-Ta Tsai.
\newblock Fast {F}ourier color constancy.
\newblock In {\em CVPR}, 2017.

\bibitem{beigpour2016multi}
Shida Beigpour, Mai~Lan Ha, Sven Kunz, Andreas Kolb, and Volker Blanz.
\newblock Multi-view multi-illuminant intrinsic dataset.
\newblock In {\em BMVC}, 2016.

\bibitem{beigpour2013multi}
Shida Beigpour, Christian Riess, Joost Van De~Weijer, and Elli Angelopoulou.
\newblock Multi-illuminant estimation with conditional random fields.
\newblock {\em IEEE Transactions on Image Processing}, 23(1):83--96, 2013.

\bibitem{bianco2019quasi}
Simone Bianco and Claudio Cusano.
\newblock Quasi-unsupervised color constancy.
\newblock In {\em CVPR}, 2019.

\bibitem{bianco2012color}
Simone Bianco and Raimondo Schettini.
\newblock Color constancy using faces.
\newblock In {\em CVPR}, 2012.

\bibitem{bleier2011color}
Michael Bleier, Christian Riess, Shida Beigpour, Eva Eibenberger, Elli
  Angelopoulou, Tobias Tr{\"o}ger, and Andr{\'e} Kaup.
\newblock Color constancy and non-uniform illumination: Can existing algorithms
  work?
\newblock In {\em ICCV Workshops}, 2011.

\bibitem{brainard1994bayesian}
David~H Brainard and William~T Freeman.
\newblock Bayesian method for recovering surface and illuminant properties from
  photosensor responses.
\newblock In {\em Human Vision, Visual Processing, and Digital Display V},
  volume 2179, pages 364--377, 1994.

\bibitem{brainard1997bayesian}
David~H Brainard and William~T Freeman.
\newblock Bayesian color constancy.
\newblock {\em Journal of the Optical Society of America A}, 14(7):1393--1411,
  1997.

\bibitem{maxRGB}
David~H Brainard and Brian~A Wandell.
\newblock Analysis of the retinex theory of color vision.
\newblock {\em Journal of the Optical Society of America A}, 3(10):1651--1661,
  1986.

\bibitem{GW}
Gershon Buchsbaum.
\newblock A spatial processor model for object colour perception.
\newblock {\em Journal of the Franklin Institute}, 310(1):1--26, 1980.

\bibitem{fivek}
Vladimir Bychkovsky, Sylvain Paris, Eric Chan, and Fr{\'e}do Durand.
\newblock Learning photographic global tonal adjustment with a database of
  input / output image pairs.
\newblock In {\em CVPR}, 2011.

\bibitem{cardei2002estimating}
Vlad~C Cardei, Brian Funt, and Kobus Barnard.
\newblock Estimating the scene illumination chromaticity by using a neural
  network.
\newblock {\em Journal of the Optical Society of America A}, 19(12):2374--2386,
  2002.

\bibitem{cheng2014illuminant}
Dongliang Cheng, Dilip~K Prasad, and Michael~S Brown.
\newblock Illuminant estimation for color constancy: {W}hy spatial-domain
  methods work and the role of the color distribution.
\newblock {\em Journal of the Optical Society of America A}, 31(5):1049--1058,
  2014.

\bibitem{ebner2007color}
Marc Ebner.
\newblock {\em Color {C}onstancy}, volume~6.
\newblock John Wiley \& Sons, 2007.

\bibitem{finlayson2000improving}
Graham Finlayson and Steven Hordley.
\newblock Improving gamut mapping color constancy.
\newblock {\em IEEE Transactions on Image Processing}, 9(10):1774--1783, 2000.

\bibitem{MomentCorrection}
Graham~D Finlayson.
\newblock Corrected-moment illuminant estimation.
\newblock In {\em ICCV}, 2013.

\bibitem{royalsociety}
Graham~D. Finlayson.
\newblock Colour and illumination in computer vision.
\newblock {\em Interface Focus}, 8(4):1--8, 2018.

\bibitem{finlayson1995color}
Graham~D Finlayson, Brian~V Funt, and Kobus Barnard.
\newblock Color constancy under varying illumination.
\newblock In {\em ICCV}, pages 720--725, 1995.

\bibitem{Gamut1}
Graham~D Finlayson, Steven~D Hordley, and Ingeborg Tastl.
\newblock Gamut constrained illuminant estimation.
\newblock {\em International Journal of Computer Vision}, 67(1):93--109, 2006.

\bibitem{SoG}
Graham~D Finlayson and Elisabetta Trezzi.
\newblock Shades of gray and colour constancy.
\newblock In {\em Color and Imaging Conference}, 2004.

\bibitem{forsyth1990novel}
David~A Forsyth.
\newblock A novel algorithm for color constancy.
\newblock {\em International Journal of Computer Vision}, 5(1):5--35, 1990.

\bibitem{fourure2017residual}
Damien Fourure, R{\'e}mi Emonet, Elisa Fromont, Damien Muselet, Alain Tremeau,
  and Christian Wolf.
\newblock Residual conv-deconv grid network for semantic segmentation.
\newblock In {\em BMVC}, 2017.

\bibitem{funt1996learning}
Brian Funt, Vlad Cardei, and Kobus Barnard.
\newblock Learning color constancy.
\newblock In {\em Color and Imaging Conference}, 1996.

\bibitem{gehler2008bayesian}
Peter~Vincent Gehler, Carsten Rother, Andrew Blake, Tom Minka, and Toby Sharp.
\newblock Bayesian color constancy revisited.
\newblock In {\em CVPR}, 2008.

\bibitem{Gamut}
Arjan Gijsenij, Theo Gevers, and Joost Van De~Weijer.
\newblock Generalized gamut mapping using image derivative structures for color
  constancy.
\newblock {\em International Journal of Computer Vision}, 86(2):127--139, 2010.

\bibitem{gijsenij2011computational}
Arjan Gijsenij, Theo Gevers, and Joost Van De~Weijer.
\newblock Computational color constancy: Survey and experiments.
\newblock {\em IEEE Transactions on Image Processing}, 20(9):2475--2489, 2011.

\bibitem{gijsenij2012improving}
Arjan Gijsenij, Theo Gevers, and Joost Van De~Weijer.
\newblock Improving color constancy by photometric edge weighting.
\newblock {\em IEEE Transactions on Pattern Analysis and Machine Intelligence},
  34(5):918--929, 2012.

\bibitem{gijsenij2011color}
Arjan Gijsenij, Rui Lu, and Theo Gevers.
\newblock Color constancy for multiple light sources.
\newblock {\em IEEE Transactions on Image Processing}, 21(2):697--707, 2011.

\bibitem{hao2020multi}
Xiangpeng Hao and Brian Funt.
\newblock A multi-illuminant synthetic image test set.
\newblock {\em Color Research \& Application}, 45(6):1055--1066, 2020.

\bibitem{hao2019evaluating}
Xiangpeng Hao, Brian Funt, and Hanxiao Jiang.
\newblock Evaluating colour constancy on the new mist dataset of
  multi-illuminant scenes.
\newblock In {\em Color and Imaging Conference}, volume 2019, 2019.

\bibitem{hernandez2020multi}
Daniel Hernandez-Juarez, Sarah Parisot, Benjamin Busam, Ales Leonardis, Gregory
  Slabaugh, and Steven McDonagh.
\newblock A multi-hypothesis approach to color constancy.
\newblock In {\em CVPR}, 2020.

\bibitem{hong2001study}
Guowei Hong, M~Ronnier Luo, and Peter~A Rhodes.
\newblock A study of digital camera colorimetric characterization based on
  polynomial modeling.
\newblock {\em Color Research \& Application}, 26(1):76--84, 2001.

\bibitem{hsu2008light}
Eugene Hsu, Tom Mertens, Sylvain Paris, Shai Avidan, and Fr{\'e}do Durand.
\newblock Light mixture estimation for spatially varying white balance.
\newblock In {\em SIGGRAPH}. 2008.

\bibitem{hu2017fc}
Yuanming Hu, Baoyuan Wang, and Stephen Lin.
\newblock {FC4}: Fully convolutional color constancy with confidence-weighted
  pooling.
\newblock In {\em CVPR}, 2017.

\bibitem{hussain2018color}
Md~Akmol Hussain and Akbar~Sheikh Akbari.
\newblock Color constancy algorithm for mixed-illuminant scene images.
\newblock {\em IEEE Access}, 6:8964--8976, 2018.

\bibitem{joze2013exemplar}
Hamid Reza~Vaezi Joze and Mark~S Drew.
\newblock Exemplar-based color constancy and multiple illumination.
\newblock {\em IEEE Transactions on Pattern Analysis and Machine Intelligence},
  36(5):860--873, 2013.

\bibitem{BP}
Hamid Reza~Vaezi Joze, Mark~S Drew, Graham~D Finlayson, and Perla
  Aurora~Troncoso Rey.
\newblock The role of bright pixels in illumination estimation.
\newblock In {\em Color and Imaging Conference}, 2012.

\bibitem{kingma2014adam}
Diederik~P Kingma and Jimmy Ba.
\newblock Adam: A method for stochastic optimization.
\newblock {\em arXiv preprint arXiv:1412.6980}, 2014.

\bibitem{lo2021clcc}
Yi-Chen Lo, Chia-Che Chang, Hsuan-Chao Chiu, Yu-Hao Huang, Chia-Ping Chen,
  Yu-Lin Chang, and Kevin Jou.
\newblock {CLCC}: Contrastive learning for color constancy.
\newblock In {\em CVPR}, 2021.

\bibitem{BMVC1}
Zhongyu Lou, Theo Gevers, Ninghang Hu, Marcel~P Lucassen, et~al.
\newblock Color constancy by deep learning.
\newblock In {\em BMVC}, 2015.

\bibitem{murmann2019dataset}
Lukas Murmann, Michael Gharbi, Miika Aittala, and Fredo Durand.
\newblock A dataset of multi-illumination images in the wild.
\newblock In {\em ICCV}, 2019.

\bibitem{niklaus2018context}
Simon Niklaus and Feng Liu.
\newblock Context-aware synthesis for video frame interpolation.
\newblock In {\em CVPR}, 2018.

\bibitem{Seoung}
Seoung~Wug Oh and Seon~Joo Kim.
\newblock Approaching the computational color constancy as a classification
  problem through deep learning.
\newblock {\em Pattern Recognition}, 61:405--416, 2017.

\bibitem{GP}
Yanlin Qian, Ke Chen, Jarno Nikkanen, Joni-Kristian K{\"a}m{\"a}r{\"a}inen, and
  Jiri Matas.
\newblock Revisiting gray pixel for statistical illumination estimation.
\newblock In {\em VISAPP}, 2019.

\bibitem{GI}
Yanlin Qian, Joni-Kristian K{\"a}m{\"a}r{\"a}inen, Jarno Nikkanen, and Jiri
  Matas.
\newblock On finding gray pixels.
\newblock In {\em CVPR}, 2019.

\bibitem{rosenberg2004bayesian}
Charles Rosenberg, Alok Ladsariya, and Tom Minka.
\newblock Bayesian color constancy with non-gaussian models.
\newblock In {\em NeurIPS}, 2004.

\bibitem{sharma2005ciede2000}
Gaurav Sharma, Wencheng Wu, and Edul~N Dalal.
\newblock The {C}{I}{E}{D}{E}2000 color-difference formula: Implementation
  notes, supplementary test data, and mathematical observations.
\newblock {\em Color Research \& Application}, 30(1):21--30, 2005.

\bibitem{DSNET}
Wu Shi, Chen~Change Loy, and Xiaoou Tang.
\newblock Deep specialized network for illuminant estimation.
\newblock In {\em ECCV}, 2016.

\bibitem{3dsmax}
Sham Tickoo.
\newblock {\em Autodesk 3{D}s Max 2021: A comprehensive guide}.
\newblock Cadcim Technologies, 2020.

\bibitem{GE}
Joost Van De~Weijer, Theo Gevers, and Arjan Gijsenij.
\newblock Edge-based color constancy.
\newblock {\em IEEE Transactions on Image Processing}, 16(9):2207--2214, 2007.

\bibitem{xu2020end}
Bolei Xu, Jingxin Liu, Xianxu Hou, Bozhi Liu, and Guoping Qiu.
\newblock End-to-end illuminant estimation based on deep metric learning.
\newblock In {\em CVPR}, 2020.

\end{thebibliography}
}

\end{document}